\documentclass[10pt,journal,compsoc]{IEEEtran}

\ifCLASSOPTIONcompsoc
\usepackage[nocompress]{cite}
\else
\usepackage{cite}
\fi

\ifCLASSINFOpdf
\usepackage[pdftex]{graphicx}
\DeclareGraphicsExtensions{.pdf,.jpeg,.png}
\DeclareGraphicsExtensions{.eps}
\usepackage{epstopdf}

\else

\fi

\ifCLASSOPTIONcompsoc
\usepackage[caption=false,font=footnotesize,labelfont=sf,textfont=sf]{subfig}
\else
\usepackage[caption=false,font=footnotesize]{subfig}
\fi
\usepackage{booktabs}
\usepackage{amssymb}

\usepackage{mathrsfs}
\usepackage{textcomp}
\usepackage{epsfig}
\usepackage{graphicx}
\usepackage{amsmath,amsfonts}
\usepackage{enumerate}
\usepackage{cases}
\usepackage{multirow}
\usepackage{verbatim}
\usepackage{amssymb}
\usepackage{CJK}
\usepackage{algorithmicx}

\usepackage{color}
\usepackage{bm}
\usepackage{booktabs}
\usepackage{colortbl}
\usepackage{float}
\usepackage{ragged2e}  
\usepackage{csquotes}
\usepackage[breaklinks=true,bookmarks=false]{hyperref}
\usepackage{diagbox}
\usepackage{threeparttable}
\usepackage{xspace}
\usepackage{array}
\usepackage[table]{xcolor}
\usepackage{overpic}
\usepackage{textcomp}
\usepackage{contour}
\usepackage[table]{xcolor}
\usepackage{threeparttable}
\usepackage{tablefootnote}

\newcommand{\tabincell}[2]{\begin{tabular}{@{}#1@{}}#2\end{tabular}}

\usepackage{silence}
\makeatletter
\newcommand{\thickhline}{%
	\noalign {\ifnum 0=`}\fi \hrule height 1pt
	\futurelet \reserved@a \@xhline
}
\makeatother

\makeatletter
\DeclareRobustCommand\onedot{\futurelet\@let@token\@onedot}
\def\@onedot{\ifx\@let@token.\else.\null\fi\xspace}

\begin{document}

	\title{A Systematic Investigation on Deep Learning-Based Omnidirectional Image and Video Super-Resolution}

	\author{Qianqian Zhao, Chunle Guo, Tianyi Zhang, Junpei Zhang, Peiyang Jia, Tan Su, Wenjie Jiang,  Chongyi Li
		
		\thanks{Q. Zhao, C. Guo, T. Zhang, and C. Li are  with the College of Computer Science, Nankai University, Tianjin, China (e-mail: qqzhao@mail.nankai.edu.cn, guochunle@nankai.edu.cn, and 2120240704@mail.nankai.edu.cn, lichongyi@nankai.edu.cn).}
		\thanks{J. Zhang, P. Jia, T. Su, and W. Jiang are with the Insta360, China  (e-mail:zhangjunpei@insta360.com, jiapeiyang@insta360.com, sutan@insta360.com, and jerett@insta360.com).}

	}
	
	\markboth{}%
	{Shell \MakeLowercase{\textit{et al.}}: Bare Demo of IEEEtran.cls for Computer Society Journals}

	\IEEEtitleabstractindextext{%
		\justify  

\begin{abstract}
	\label{sec:Abstrat}
Omnidirectional image and video super-resolution is a crucial research topic in low-level vision, playing an essential role in virtual reality and augmented reality applications. 
Its goal is to reconstruct high-resolution images or video frames from low-resolution inputs, thereby enhancing detail preservation and enabling more accurate scene analysis and interpretation.
In recent years, numerous innovative and effective approaches have been proposed, predominantly based on deep learning techniques, involving diverse network architectures, loss functions, projection strategies, and training datasets.
This paper presents a systematic review of recent progress in omnidirectional image and video super-resolution, focusing on deep learning-based methods. 
Given that existing datasets predominantly rely on synthetic degradation and fall short in capturing real-world distortions, we introduce a new dataset, 360Insta, that comprises authentically degraded omnidirectional images and videos collected under diverse conditions, including varying lighting, motion, and exposure settings.
This dataset addresses a critical gap in current omnidirectional benchmarks and enables more robust evaluation of the generalization capabilities of omnidirectional super-resolution methods.
We conduct comprehensive qualitative and quantitative evaluations of existing methods on both public datasets and our proposed dataset. 
Furthermore, we provide a systematic overview of the current status of research and discuss promising directions for future exploration. 
All datasets, methods, and evaluation metrics introduced in this work are publicly available and will be regularly updated.
Project page: \url{https://github.com/nqian1/Survey-on-ODISR-and-ODVSR}.

\end{abstract}

		}
	
	\maketitle

	\IEEEdisplaynontitleabstractindextext
	
	\IEEEpeerreviewmaketitle


\IEEEraisesectionheading{\section{Introduction}
	\label{sec:Introduction}}
\IEEEPARstart{F}{ollowing} the rapid advancement of information technology, Virtual Reality (VR) and Augmented Reality (AR) have gained increasing attention in both academia and industry.
Omnidirectional images (ODIs), commonly referred to as ODIs, provide a 360°$\times$180° field of view (FoV) and have become foundational for VR, AR, and metaverse. They offer users a comprehensive perspective and full interactive capabilities, facilitating an immersive experience.

\begin{figure*}[t]  
    \centering  
    \includegraphics[width=0.98\textwidth]{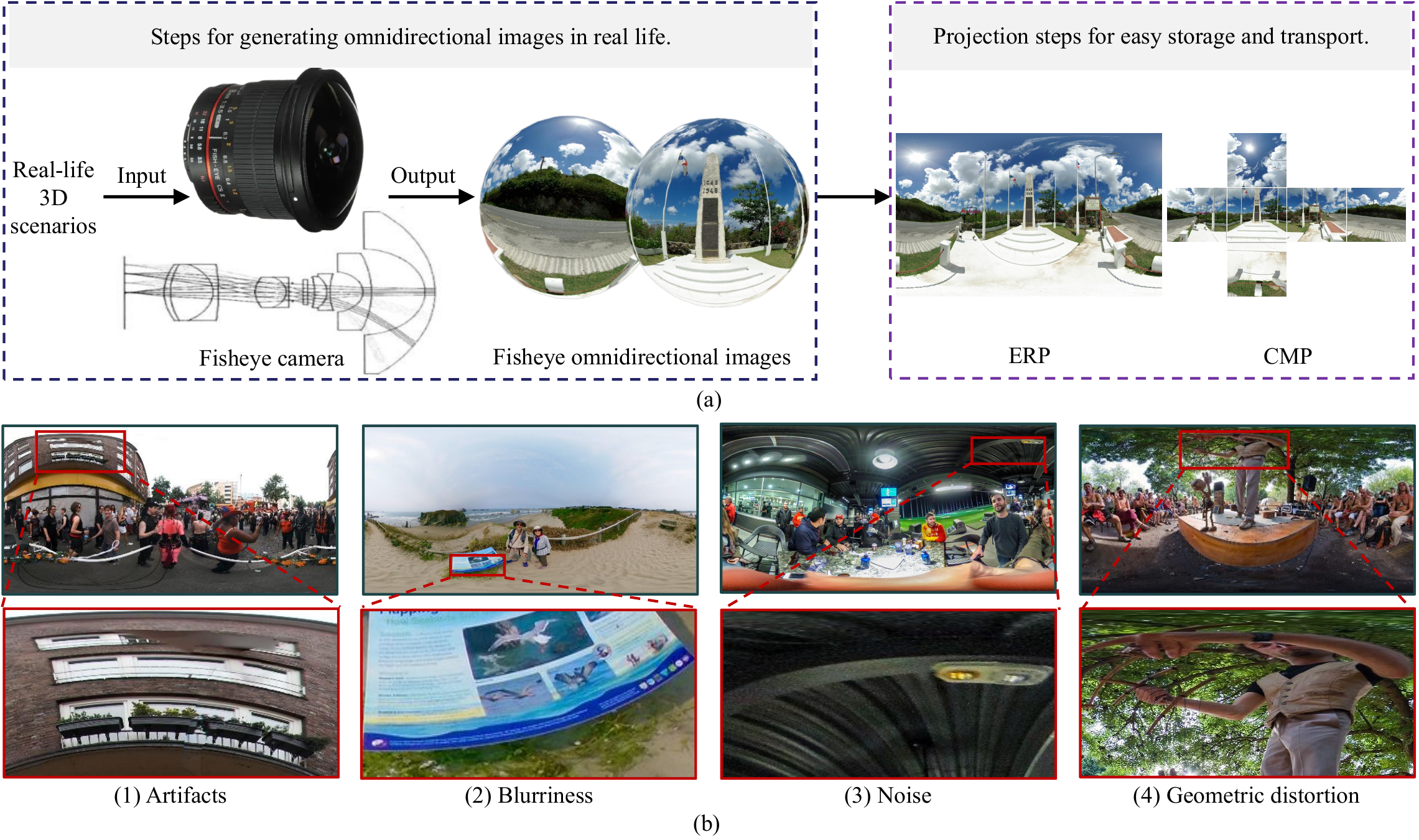}   
    \caption{Overview of omnidirectional imaging. (a) Omnidirectional imaging and projection. (b) Examples of LR ODIs captured in the real world.}  
    \label{fig:process}  
\end{figure*}

As illustrated in Fig.~\ref{fig:process}(a), fisheye cameras capture real-world scenes to generate fisheye ODIs. 
Then, low-resolution (LR) equidistant rectangular projection (ERP) images or cubemap projection (CMP) are commonly stored and transmitted to optimize memory usage and bandwidth. 
However, as shown in Fig.~\ref{fig:process}(b), these LR ERP images are often affected by artifacts, blurriness, noise, and geometric distortions, all of which significantly impair the user's immersive experience. 
Consequently, capturing and transmitting high-resolution (HR) ODIs is essential to ensure visual quality and user satisfaction in immersive applications.

Omnidirectional image super-resolution (ODISR) aims to generate HR images or video frames from LR images or video frames, thereby revealing finer details and enabling more accurate scene analysis and interpretation.
The early methods include a variety of techniques, such as regularization methods, neighborhood embedding-based algorithms, projection onto convex sets (POCS), and algorithms utilizing redundant similar blocks in LR images to reconstruct HR images~\cite{low,single,image,superosr,pocs}. 
In addition, some studies on ODIs introduce innovative SR methods, using ODIs from hand-crafted methods or simple learning-based methods. 
For example, in~\cite{earlyplenoptic}, the authors applied plenoptic geometry to register consecutive frames of an ODV. 
This process allows them to use the aggregated visual information to generate HR ODIs. 
Later, Arican et al.~\cite{earlyjoint,superosr} demonstrated how multiple ODIs with arbitrary rotations could be used to solve the SR challenge by exploiting the spherical fourier transform (SFT). 
In their work, the joint registration and SR challenge is solved through the minimization of the total least squares norm in the SFT domain.
However, these traditional algorithms often struggle with large-scale image magnification, making it difficult to meet the performance requirements for high-quality reconstruction.

In recent years, deep learning-based methods have rapidly advanced, significantly contributing to single-image super-resolution (SISR). 
Following the initial application of SR methods based on convolutional neural networks (CNN)~\cite{cnnsr,drcnn,dccn,mssr}, algorithms based on generative adversarial networks (GAN)~\cite{mlsr,calgan,noise}, vision transformers (ViT)~\cite{srformer,msra,apisr}, and diffusion models~\cite{ecdmsr,sinsr,acdmsr,adadiffsr} have further advanced this technology. 
Among these, SwinIR~\cite{swinsr}, a Swin Transformer-based image super-resolution (ISR) method, effectively addresses long-range dependency issues of images using the hierarchical structure and local attention mechanism of the Swin Transformer, demonstrating superior performance in SISR tasks. 
However, traditional SR methods face significant challenges when applied to ODIs. 
Specifically, the severe distortion of ERP in high-latitude areas and wide viewing angles causes geometric deformations and pixel distortions, making it difficult to directly apply existing methods. 
Furthermore, the large FoV of ODI data introduces discontinuities between pixels, which can lead to splicing errors and misalignment, particularly at the edges. 
These factors complicate the pixel relationships, and traditional SR algorithms struggle to accurately restore details, failing to effectively handle the challenges posed by distortion and misalignment.

\begin{figure*}[t!]
	\centering  \centerline{\includegraphics[width=1\linewidth]{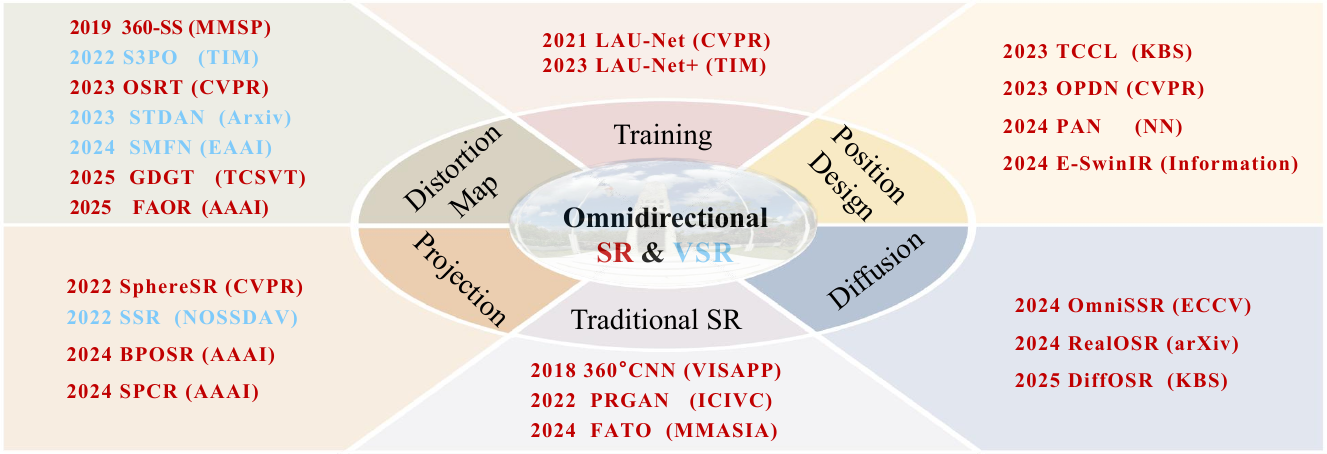}}
	\vspace{-2pt}
	\caption{A review of deep learning-based {\color{red}ODISR} ({\color{red}red}) and {\color[HTML]{87CEFA}ODVSR} ({\color[HTML]{87CEFA}blue}) methods. The Distortion Map~\cite{360SS, S3PO,OSRT,FAOR,GRGTN,SMFN} refers to the fact that the distortion map is used as a condition to guide model training. Projection~\cite{spheresr,VertexShuffle,BPOSR,SPCR} refers to that the model is designed using multiple projection methods such as hexahedron and cylindrical projection. Training~\cite{lau,lau+} refers to enhancing SR performance through optimized training strategies. Traditional SR~\cite{360CNN,PRGAN,FATO} indicates that only the traditional SR structure is used to perform the ODISR task. Position
    Design~\cite{TCCL,OPDN} involves adjustments based on encoding absolute or relative positions within ODIs. Diffusion~\cite{omnissr,realosr,diffosr} refers to that diffusion models are used to address the task of ODISR.}
	\label{fig:milestones}
	\vspace{-4pt}
\end{figure*}

To address these unique challenges, methods specifically designed for ODISR have been proposed, with a particular focus on deep learning-based methods. Fig.~\ref{fig:milestones} shows a concise overview of the deep learning-based ODISR and ODVSR methods. 
As shown, since 2018, the number of deep learning-based ODISR solutions has gradually become mainstream. These methods are based on the traditional SR network structure and combine the characteristics of ODISR and ODVSR to make innovative adjustments, which can be roughly summarized as follows:
Distortion Map~\cite{OSRT,POOISR,OPDN,GDGT}, Projection~\cite{BPOSR,SPCR}, Training~\cite{lau,lau+}, SR~\cite{360CNN}, Position Design~\cite{TCCL,OPDN} and Diffusion~\cite{omnissr,realosr,diffosr}.
Furthermore, Cao et al.~\cite{2023ntire} and Telili et al.~\cite{360MR} organized respective competitions focused on ODISR and ODVSR. Cao et al.~\cite{2023ntire} primarily introduced two high-quality data sets: Flickr360~\cite{2023ntire} and ODV360~\cite{2023ntire} for the ODISR and ODVSR task, and provided comparisons and analysis of the top-performing models of the challenge. In addition, Telili et al.~\cite{360MR} curated a benchmark for ODVSR and quality enhancement, evaluating state-of-the-art methods within a unified framework that considers quality improvement, bitrate savings, and computational efficiency. However, both works lack a detailed classification and review of the latest research in the ODISR and ODVSR tasks. Moreover, the datasets used in these competitions are based on synthetic bicubic downsampling, which does not account for the complexities of real-world degradations.

\begin{figure}[t]  
    \centering  
\includegraphics[scale=0.57]{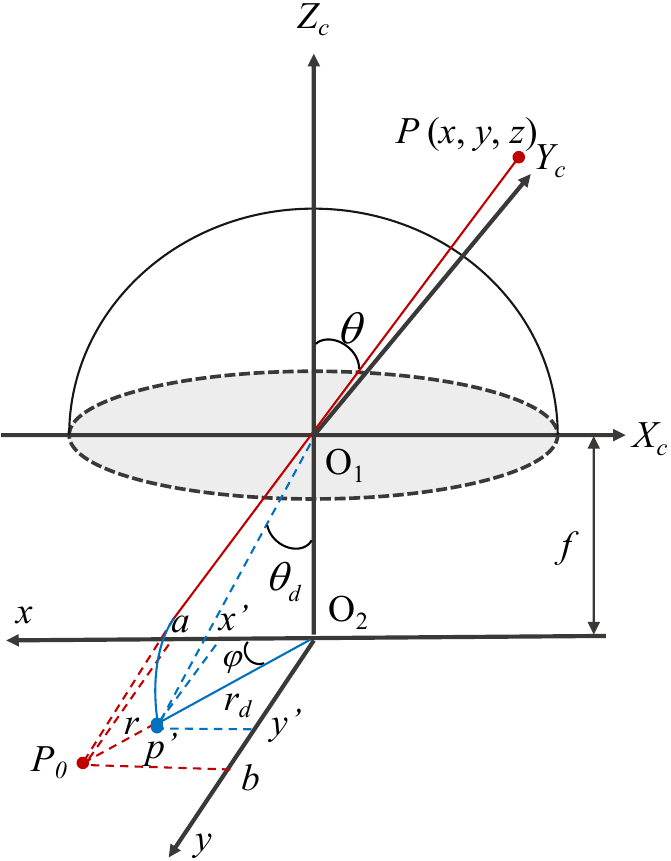}   
    \caption{Fisheye camera imaging principle. Spherical Projection: 3D points on the normalized plane are linearly projected onto a virtual unit sphere centered at the camera coordinate system origin. Nonlinear Projection: The points on the sphere are then projected onto the image plane. This process is nonlinear, resulting in the characteristic fisheye distortion.}  
    \label{fig:fisheye}  
\end{figure} 


To tackle the above problems, our work primarily focuses on the latest advancements in deep learning-based ODISR and ODVSR. We conduct in-depth analysis and discussion from various aspects, including network architectures, common projections, training datasets, testing datasets, evaluation metrics, model sizes, and inference speeds. In addition, our review has the following features.

1) We have organized the currently testable ODISR and ODVSR models and unified the current messy test data and test protocols. Organizing test data and protocols provides a reliable and reproducible evaluation platform for subsequent research and improves the transparency of research.

2) We propose a new 360Insta dataset to address a key gap in current research. Existing omnidirectional datasets are predominantly synthetic or collected from the Internet, often involving artificially simulated degradations that do not fully capture real-world conditions. In contrast, our 360Insta dataset is captured using a 360 Insta omnidirectional camera in real environments, encompassing diverse scene types such as indoor, outdoor, daytime, nighttime, and scenes with people. Importantly, we also include images with various real-world degradations, such as varying lighting conditions, motion blur, dim conditions, and different exposures, to better evaluate the robustness of existing algorithms in practical and challenging scenarios. This makes our dataset a valuable complement to existing synthetic datasets, which may not fully reflect real-world distortions, thereby providing a more rigorous benchmark for current methods. To the best of our knowledge, this is the first work to compare the performance of multiple deep learning-based omnidirectional image processing techniques on such real-world degraded data, enabling more accurate assessments of their practical applicability.

3) To promote the continued development of this field, we maintain and regularly update the GitHub website, which discloses the latest research progress and covers many deep learning-based ODISR and ODVSR methods. At the same time, we also provide download links for related datasets to facilitate researchers in obtaining data for experiments. This initiative accelerates research communication and innovation and helps inspire new research directions.

We hope that the survey offers new insights and inspiration to deepen the understanding of deep learning-based ODISR, promote further research on the identified open challenges, and accelerate progress in this evolving field.

\section{Fisheye Imaging Projection Principle}
\label{sec:fisheye}

In this section, we will introduce the principle of omnidirectional imaging, namely the fisheye imaging projection principle~\cite{fisheye1,fisheye2}. The above imaging principle will provide theoretical support for the degradation design and the method proposal of subsequent omnidirectional data sets.

The imaging of the fisheye camera is illustrated in Fig.~\ref{fig:fisheye}. The camera coordinate system is denoted by $O_1-X_cY_cZ_c$, and the imaging plane coordinate system is denoted by $O_2-xy$. $f$ denotes the focal length of the fisheye camera, typically set to 1. Consider a point $P$ in the real world with an incident angle $\theta$. Under the imaging model of an ordinary pinhole camera, $P$, $O_1$, and $P_0$ are collinear, where $P_0$ denotes the image of $P$. Due to the optical design of the fisheye lens, which introduces geometric distortion, the final imaging point is $p'$, with its position described in polar coordinates as $(r_d, \varphi)$. The detailed process of the fisheye image formation is as follows.

Given a point $P = (x, y, z)$ in the three-dimensional space that represents a location in the real world, its projection onto the image plane under the pinhole camera model, assuming no distortion, results in an image point $P_0 = (a, b)$. The coordinates of $P_0$ and the incident angle $\theta$ are computed according to the following equations:
\begin{equation}  
a = \frac{x}{z}, \quad b = \frac{y}{z},
\end{equation}  
\begin{equation}  
r = \left| O_2P_0 \right| = \sqrt{a^2 + b^2},
\end{equation}  
\begin{equation}  
\theta = \arctan(r),  
\end{equation}  

To project the largest possible scene into a limited image plane, fisheye cameras are designed according to certain projection functions (Perspective projection, Level projection, Equirectangular projection, Hint projection, and Orthogonal projection). However, real lenses do not precisely conform to the ideal projection model. To facilitate fisheye camera calibration,
Kannala et al.~\cite{fisheye1} proposed a general polynomial approximation model for fisheye cameras. Analysis of the previous projection functions indicates that $\theta_d$ is an odd function of $\theta$. Furthermore, by expanding these models into Taylor series, it can be shown that $\theta_d$ can be expressed as a polynomial that involves only odd powers of $\theta$. For that, it is common to approximate the actual fisheye projection function by taking the first five terms of the Taylor series expansion of $r$ with respect to $\theta$.

In addition, due to the existence of distortion, the distance $r$ from the image point to $O_2$ is compressed to $r_d$, and the actual position of the image point is $p'(x', y')$. The $r_d$ is calculated as follows:
\begin{equation}
r_d = \left| O_2P' \right| = \theta_d = \theta + k_1 \theta^3 + k_2 \theta^5 + k_3 \theta^7 + k_4 \theta^9,
\end{equation} 

Then, the actual imaging point $p'(x',y')$ is calculated as:
\begin{equation}  
x' = \frac{\theta_d}{r} a, \quad y' = \frac{\theta_d}{r} b.  
\end{equation}  

Finally, $p'$ is converted to the pixel coordinate system $(u, v)$ according to the intrinsic parameters of the camera:
\begin{equation}  
\begin{aligned}  
u &= f_x x' + c_x, \\
v &= f_y y' + c_y,  
\end{aligned}  
\end{equation}  
where $f_x$ and $f_y$ represent the focal length and are used to transform the points on the image into the camera coordinate system.
$(c_x, c_y)$ represents the center point of the image, expressed as pixel values. It is the offset that corrects the distorted point to the actual pixel coordinate system.


\section{Deep Learning-Based ODISR}

	\begin{table*}
		\rowcolors{1}{gray!20}{white}
		\centering
		\caption{
			{Summary of essential characteristics of representative deep learning-based ODISR methods.}
		}
		\vspace{-6pt}
		\label{table:odisrmethods}
		\begin{threeparttable}
			\resizebox{1\textwidth}{!}{
				\setlength\tabcolsep{2pt}
				\renewcommand\arraystretch{0.96}
                \centering
				\begin{tabular}{c|c||c|c|c|c|c}
					\hline

					&\textbf{Method}&\textbf{Classification} &\textbf{Loss Function}
					&\textbf{Training Data} & \textbf{Testing Data} &\textbf{Evaluation Metric}  \\
					\hline
					\hline
					\multirow{1}{*}{\rotatebox{90}{\textbf{2018}}}
					&360°CNN~\cite{360CNN}  &\tabincell{c}{Traditional SR} &MSE Loss
					&\tabincell{c}{Collection \\ of \\ Nokia OZO\\ } &\tabincell{c}{Collection \\ of\\ Nokia OZO} &\tabincell{c}{PSNR}  \\
					\hline
					\hline
                    
					\multirow{1}{*}{\rotatebox{90}{\textbf{2019}}}
					&360-SS~\cite{360SS} 
					&\tabincell{c}{Distortion Map} &\tabincell{c}{360-SS Loss\\GAN Loss Content Loss} &\tabincell{c}{Randomly selected a total of\\ 4500 ODIs from
                   the \\SUN 360 Database} &\tabincell{c}{Randomly selected a total of\\ 500 ODIs from the \\SUN 360 Database} &\tabincell{c}{SSIM PSNR \\WS-SSIM WS-PSNR}\\
                        \hline
					\hline
                    
                    \multirow{1}{*}{\rotatebox{90}{\textbf{2021}}}
					&360SISR~\cite{360SISR} &Distortion Map
					&\tabincell{c}{WS-L1 Loss} &\tabincell{c}{DIV2K\\ DistortedDIV2K \\YouTube360} &Panoramic Object Detection Dataset &\tabincell{c}{SSIM PSNR \\WS-SSIM WS-PSNR} \\
					&LAU-Net~\cite{lau}  & Training &WS-L1 Loss &ODISR& \tabincell{c}{ODISR \\100 ODIs from the \\SUN 360 Database} &WS-SSIM WS-PSNR \\
                        \hline
					\hline
                    
                    \multirow{1}{*}{\rotatebox{90}{\textbf{2022}}}
					&360PRGAN~\cite{PRGAN} &\tabincell{c}{Traditional SR}
					&\tabincell{c}{MSE Loss\\VGG Loss\\GAN Loss} &360SP &360SP &\tabincell{c}{VIFP\\SSIM PSNR}\\
					&SphereSR~\cite{spheresr}  &Projection
					&\tabincell{c}{L1 Loss \\Feature Loss} &ODISR& \tabincell{c}{ODISR \\100 ODIs from the \\SUN 360 Database} &\tabincell{c}{SSIM PSNR \\WS-SSIM WS-PSNR}\\
                        \hline
					\hline
                    
                    \multirow{1}{*}{\rotatebox{90}{\textbf{2023}}}
                    
					&LAU-Net+~\cite{lau+}  & Training &\tabincell{c}{WS-L1 Loss \\L1 Loss }&ODISR& \tabincell{c}{ODISR \\100 ODIs from the \\SUN 360 Database} &WS-SSIM WS-PSNR \\
                    
                    &OSRT~\cite{OSRT}  & Distortion Map &L1 Loss &ODISR& \tabincell{c}{ODISR \\100 ODIs from the \\SUN 360 Database} &\tabincell{c}{SSIM PSNR \\WS-SSIM WS-PSNR}\\
    			&POOISR~\cite{POOISR}  &  Distortion Map &Perception-orientated Adaptive Loss &ODISR& \tabincell{c}{ODISR \\100 ODIs from the \\SUN 360 Database} &\tabincell{c}{SSIM PSNR \\WS-SSIM WS-PSNR}\\
                    &OPDN~\cite{OPDN}  &  Position Design &L2 Loss &\tabincell{c}{Flickr360 dataset\\Collection an extra 260 HR \\360 videos from
         YouTube}& Flickr360 &\tabincell{c}WS-SSIM WS-PSNR\\
                     &TCCL~\cite{TCCL}  & Position Design &\tabincell{c}{Pixel Loss\\ L1 Loss }&ODISR& \tabincell{c}{ODISR \\100 ODIs from the \\SUN 360 Database} &WS-SSIM WS-PSNR \\
					\hline
					\hline
					\multirow{1}{*}{\rotatebox{90}{\textbf{2024}}}
					&E-SwinIR~\cite{LTMSwinIR}&Position Design
					&L1 Loss &ODISR& \tabincell{c}{ODISR \\100 ODIs from the\\ SUN 360 Database} &\tabincell{c}{SSIM PSNR \\WS-SSIM WS-PSNR}\\
					&FATO~\cite{FATO} 
					&Traditional SR &\tabincell{c}{Charbonnier Loss\\ DCT Loss HF Loss  } &ODISR& \tabincell{c}{ODISR \\100 ODIs from the \\SUN 360 Database} &\tabincell{c}{SSIM PSNR \\WS-SSIM WS-PSNR}\\
					&BPOSR~\cite{BPOSR} 
					&Projection &L1 Loss &ODISR& \tabincell{c}{ODISR \\100 ODIs from the\\ SUN 360 Database} &\tabincell{c}{WS-SSIM WS-PSNR}\\
					&PAN~\cite{PAN}&Position Design
					&WMSE Loss &ODISR& \tabincell{c}{ODISR \\100 ODIs from the \\SUN 360 Database} &\tabincell{c}{WS-SSIM WS-PSNR}\\
					&SPCR~\cite{SPCR} 
					&Projection&Viewport-based Loss&ODISR& \tabincell{c}{ODISR \\100 ODIs from the \\SUN 360 Database} &\tabincell{c}{WS-SSIM WS-PSNR}\\
					&OmniSSR~\cite{omnissr}  &Diffusion &Information Loss
					 &ODISR& \tabincell{c}{ODISR \\100 ODIs from the \\SUN 360 Database} &\tabincell{c}{WS-PSNR  WS-SSIM\\ FID LPIPS\\NIQE DISTS}\\
					&RealOSR~\cite{realosr}  &Diffusion &\tabincell{c}{L1 Loss\\ Charbonnier
                    Loss\\LPIPS Loss \\GAN Loss}
					 &ODISR& \tabincell{c}{ODISR \\100 ODIs from the \\SUN 360 Database} &\tabincell{c}{WS-PSNR  WS-SSIM\\LPIPS DISTS \\ FID NIQE \\ MUSIQ MANIQA CLIPIQA}\\
					\hline
					\hline
					\multirow{1}{*}{\rotatebox{90}{\textbf{2025}}}
					&GDGT~\cite{GDGT}  &Position Design &WS-L1 Loss &ODISR& \tabincell{c}{ODISR \\100 ODIs from the \\SUN 360 Database} &\tabincell{c}{SSIM PSNR \\WS-SSIM WS-PSNR}\\

					&FAOR~\cite{FAOR}  &Distortion Map &L1 Loss &ODISR& \tabincell{c}{ODISR \\100 ODIs from the \\SUN 360 Database} &\tabincell{c}{WS-SSIM WS-PSNR}\\
                    
					&DiffOSR~\cite{diffosr}  &Diffusion &WS-L1 Loss &ODISR& \tabincell{c}{ODISR \\100 ODIs from the \\SUN 360 Database} &\tabincell{c}{WS-SSIM WS-PSNR\\LPIPS FID}\\
                    
					&GRGTN~\cite{GRGTN}  &Distortion Map &\tabincell{c}{L1 Loss\\SSIM Loss} &ODISR& \tabincell{c}{ODISR \\100 ODIs from the \\SUN 360 Database} &\tabincell{c}{WS-SSIM WS-PSNR}\\
                    
					&MambaOSR~\cite{mambaosr}  &Distortion Map &\tabincell{c}{L1 Loss} &ODISR& \tabincell{c}{ODISR \\100 ODIs from the \\SUN 360 Database} &\tabincell{c}{WS-SSIM WS-PSNR}\\
                    
					\hline
					\hline
				\end{tabular}
			}
		\end{threeparttable}
	\end{table*}

\begin{figure*}[t]  
    \centering  
    \includegraphics[width=0.98\textwidth]{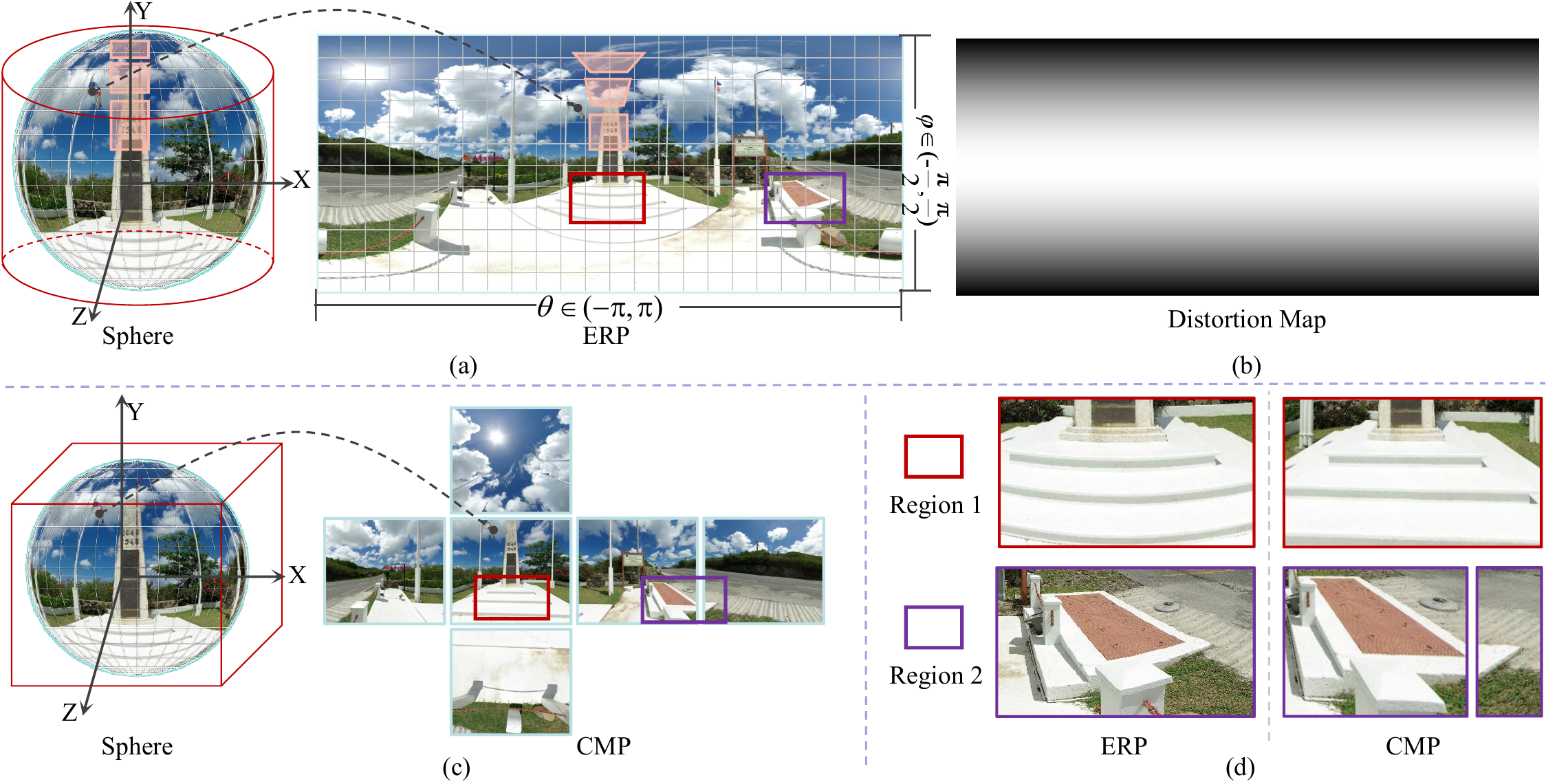}   
    \caption{ERP and CMP projections. (a) Example of spherical projection to ERP. (b) Distortion map of ERP. The increased intensity of color indicates a higher degree of distortion. (c) Example of spherical projection to CMP. (d) Comparison of selected areas between ERP and CMP. ERP facilitates seamless omnidirectional view synthesis by preserving angular continuity; however, it suffers from pronounced stretching distortions at the image periphery. Conversely, CMP minimizes spatial distortions,  but it lacks the continuity for smooth omnidirectional image. }  
    \label{fig:projection}  
\end{figure*}  

In this section, we introduce the challenges addressed by ODISR, along with the corresponding methods and their classifications, as summarized in Table~\ref{table:odisrmethods}.

\subsection{Problem Definition}
We first provide a general formulation of the ODISR problem. For a LR ODI with width $W$ and height $H$, $I_{LR} \in \mathbb{R}^{W \times H \times 3}$, the objective can be expressed as:  
\begin{align}  
I_{SR} = f(I_{LR}, \theta),   
\end{align}  
where $I_{SR} \in \mathbb{R}^{sW \times sH \times 3}$ denotes the HR image to be recovered; $s$ denotes the scale factor; $I_{LR}$ denotes the input LR image; $f$ denotes a mapping function, typically approximated by a deep learning model, with parameters $\theta$. The ultimate goal is to find the optimal network parameters that minimize the training loss: 
\begin{align}   
\hat{\theta} = \arg \min_{\theta} \mathcal{L}(I_{SR}, I_{HR}),  
\end{align}   
where $I_{HR} \in \mathbb{R}^{sW \times sH \times 3}$ is the HR target image; the loss function $\mathcal{L}(I_{SR}, I_{HR})$ represents the loss function used for optimization during the network training process.

\subsection{Traditional SR-Based ODISR Methods}
Previous methods mainly utilize traditional SR-based ODISR methods. For example, Fakour-Sevom et al.~\cite{360CNN} employ a single-image super-resolution CNN (SRCNN)~\cite{ISRCNN}, optimizing input patch size and training procedures to enhance the SR model of ODIs. 
Subsequently, Qian et al.~\cite{PRGAN} propose a Progressive Residual GAN (PRGAN) method, which uses patches from a single ERP to augment training data, incorporating residual blocks and multi-scale skip connections to progressively generate HR ODIs.
However, these methods mainly focus on data augmentation and architectural improvements, neglecting structural characteristics of ODIs, such as distortions.

\subsection{Distortion Map-Guided ODISR Methods}
With the development of immersive video, researchers have increasingly focused on the structural feature of ODIs, mainly characterized by geometric distortions. 
This is because spherical images are projected onto the ERP for transmission (Fig.~\ref{fig:process}(a)), and the uniform sampling strategy employed by the ERP causes an increase in sampling density near the poles, resulting in geometric distortions in the ERP representation.
Therefore, distortion map-guided ODISR methods primarily take into account the distortions inherent in the ERP, employing distortion maps in the design of loss functions and network architectures. 
The subsequent sections explore geometric distortion and the distortion map-guided ODISR methods (Weighted loss function design and distortion map-guided network structure design).

\subsubsection{Geometric Distortion}
As shown in Fig.~\ref{fig:projection}(a), projecting the spherical ODI to the ERP image results in distortion of the latter. This distortion varies with latitude and is symmetric across the two hemispheres. According to Equation~\eqref{eq:distor}, given a LR image $I_{LR} \in \mathbb{R}^{H \times W \times C_{in}}$, the corresponding distortion map $D \in \mathbb{R}^{H \times W \times 1}$ is defined as:  

\begin{equation} \label{eq:distor}   
D(h, 1 : W) = \cos \left( \frac{h + 0.5 - \frac{H}{2}}{H} \pi \right),   
\end{equation}  
where $D(h, 1 : W)$ represents the pixel stretching ratio from the ideal sphere to the ERP image at the current height $h$~\cite{WSPSNR}. As illustrated in Fig.~\ref{fig:projection}(b), the ERP ODIs exhibit geometric distortion and pixel stretching across latitudes, leading to substantial redundant information in high-latitude regions, which poses a challenge for current ODISR tasks.

\subsubsection{Weighted Loss Function Design}  

Recent approaches try to address the geometric distortion characteristics in ODISR. Unlike traditional SR methods that assign uniform region-based weights, ODISR utilizes pixel-wise weighting to more effectively handle spherical distortions. 
Ozcinar et al.~\cite{360SS} propose the 360-SS method, which incorporates the Weighted Structural Similarity Index (WS-SSIM) during training to emphasize the geometric positions of pixels in the spherical projection, thereby enhancing sensitivity to spherical details. 
Nishiyama et al.~\cite{360SISR} introduce a distortion-aware layer combined with a weighted L1 loss and the Weighted Peak Signal-to-Noise Ratio (WS-PSNR) during training to improve distortion handling. However, these approaches primarily treat ERP as a standard 2D image, and aside from the specialized loss functions, their network architectures remain similar to traditional SR methods, lacking explicit integration of ODI-specific features.

\subsubsection{Distortion Map-Guided Network Structure Design}  
To better utilize the information contained in the distortion map, some researchers have made improvements to network designs. 
Yu et al.~\cite{OSRT} improve the Swin Transformer by developing a distortion-aware attention module that incorporates the distortion map as a prior into the offset network, capturing the correlation between distortion and image features.
Similarly, An et al.~\cite{POOISR} propose a perception-guided adaptive loss function that jointly employs saliency and distortion maps to enhance sensitivity to distortion details. 

In summary, distortion map-guided ODISR methods incorporate the geometric distortion properties of ERP, enabling end-to-end training. However, these methods remain limited by ERP distortion characteristics, which do not resolve inherent positional discrepancies between latitudes~\cite{OPDN}. For instance, they exhibit suboptimal restoration at the left-right ERP boundary seams and fail to preserve spherical connectivity, critically undermining the integrity of the reconstructed ODIs.

\subsection{Position-Based ODISR Methods}  
Using explicit positional information in ODISR effectively addresses geometric distortions and enhances spatial awareness, thereby significantly improving the fidelity of reconstruction.

Sun et al.~\cite{OPDN} utilize sine positional encoding~\cite{attentionfor} and distortion maps~\cite{WSPSNR} to encode spherical positions, transmitting positional information to the offset network to enable deformable convolutions to adapt kernel functions based on spherical correlations, thereby enlarging receptive fields and improving reconstruction quality. 
Yao et al.~\cite{LTMSwinIR} propose a positional information transformation method that significantly improves the traditional SwinIR framework by integrating a convolution-supported deformable positional transformation module. This effectively optimizes the preprocessing stage and reduces geometric distortions, particularly in polar regions. 
Chai et al.~\cite{TCCL} argue that variations in pixel distribution across latitudes are critical factors in the design of the ODISR. They introduce absolute positional encoding into the Swin Transformer to better represent positional weights, improving the understanding of spatial relationships. Their model benefits from dual transformer and residual convolutional blocks, with a cross-guided attention mechanism that fuses long- and short-term features and facilitates reconstruction. 
Wang et al.~\cite{PAN} develop the Position Attention Network (PAN), which employs a primary enhancement branch and a position attention branch to dynamically modulate ERP features and adapt distortion characteristics.

These methods demonstrate that integrating positional encoding and deformable features is crucial to addressing distortions and improving the fidelity of reconstructed ODIs. However, existing methods remain limited by the geometric constraints of projection schemes, making it challenging to fully eliminate global spatial discrepancies.

\subsection{Projection Design-Based ODISR Methods}  
The aforementioned method utilizes ERP as input for network training; however, the ERP suffers from distortion issues. The current research aims to take advantage of the complementary advantages of various projections. In the following sections, we introduce the commonly used ERP projection and CMP projection, as well as the projection-based ODISR methods.  

\subsubsection{Image Projection Analysis} \label{sec:projection}  

As shown in Fig.~\ref{fig:projection}(a), the ERP projection process is straightforward and easy to generate, with minimal distortion in the equatorial region, making it the most commonly used projection format. 
Each coordinate $(\theta, \varphi)$ on the ideal sphere corresponds to a point $(x, y)$ on the ERP projection plane.
$(\theta, \varphi)$ are the spherical coordinates, where $(\theta, \varphi) \in [-\pi, \pi] \times \left[-\frac{\pi}{2}, \frac{\pi}{2}\right]$. The angular position $(\theta, \varphi)$ is converted to coordinates on the unit sphere $\mathbf{Q}_s = (q_{s_x}, q_{s_y}, q_{s_z})$, where
\begin{align}  
q_{s_x} &= \sin(\varphi) \cos(\theta), \\
q_{s_y} &= \sin(\theta), \\
q_{s_z} &= \cos(\varphi) \cos(\theta).  
\end{align}  
 The projection relationship~\cite{GDGT} between $(\theta, \varphi)$ and $(x, y)$ in ERP projection can be expressed as:  
\begin{equation}  
x = f(\theta, \varphi) = \theta, \quad y = h(\theta, \varphi) = \varphi.  
\end{equation}  
where $f(\cdot)$ and $h(\cdot)$ are the coordinate transformation functions from the sphere to the ERP projection plane.  

As shown in Fig.~\ref{fig:projection}(c), CMP involves projecting a sphere onto the six faces of an enclosing cube, which are then unfolded into a compact layout to form the projection plane. 
Specifically, CMP maps the sphere onto six square faces, each corresponding to a different viewing direction: front, back, left, right, top, and bottom. Each face measures $r \times r$, with a focal length of $\frac{r}{2}$, where $r$ denotes the radius of the source sphere. The front face shares the same coordinate system as the sphere, while the other faces are obtained by rotations of 90 ° or 180 ° around specific axes~\cite{cmp}.

Let $R_i$ denote the rotation matrix that transforms the coordinates from the $i$-th face's coordinate system to the spherical coordinate system. The projection of a pixel $P_c = (p^x_c, p^y_c, p^z_c)$ onto the sphere to $Q_s$ can then be described by the following transformations:  

\begin{equation}  
Q_s = s \cdot R_i \cdot P_c,  
\end{equation}  
where $ p^x_c, p^y_c \in [0, r] $, $ p^z_c = \frac{r}{2} $, and the factor $ s = \frac{r}{|p_c|} $.  

As shown in Fig.~\ref{fig:projection}(d), compared to ERP, CMP significantly reduces image distortion. However, it introduces a continuity challenge by disrupting the continuity of objects across the boundaries between different faces.

\subsubsection{Projection-Based ODISR Methods} \label{sec:projection_method}  
Significant advancements have been made in addressing the challenges of ODISR with various projection schemes.

Yoon et al.~\cite{spheresr} introduce a projection-based method, SphereSR, which models HR image reconstruction through a continuous spherical representation. By extracting features from icosahedral spherical data and employing a local spherical implicit image function, SphereSR could predict RGB values at arbitrary spherical coordinates and accommodating multiple projection types. However, reliance on multiple dedicated networks for different projections increases computational complexity, especially given the unstructured and intricate nature of polyhedral projections, which hampers scalability and efficiency.
Wang et al.~\cite{BPOSR} recognize the complementary strengths of the ERP and CMP projections, that is, full field of view versus low distortion. They propose a dual-branch network architecture that enables simultaneous feature extraction and interaction across both projections. By integrating specialized transformers for each projection, along with a fusion module that adaptively combines features based on attention mechanisms, their approach enhances the overall robustness and leverages the geometric characteristics of each projection type.
Cai et al.~\cite{SPCR} propose an innovative spherical pseudo-cylindrical representation, which adaptively adjusts sampling density across latitudes. This method addresses the geometric distortion issues inherent in ERP projections while maintaining compatibility with existing SR models.

In summary, recent research has attempted to utilize the advantages of ERP and CMP for HR image reconstruction. However, challenges such as computational complexity and dependence on input quality remain. 

\subsection{Training-Based ODISR Methods} 
Deng et al. introduced LAU-Net~\cite{lau} and LAU-Net+~\cite{lau+} as methods that leverage latitude-aware structures to adaptively handle the diverse pixel density and texture complexity across latitude bands in ODISR.

In training, LAU-Net~\cite{lau} adopts a multi-level pyramid architecture, where each level's parameters are optimized to extract high-level features via attention mechanisms and spatial partitioning. The dynamic band pruning process at each stage is guided by learned criteria, aiming to improve the effectiveness of upsampling factors.
LAU-Net+~\cite{lau+} extends this by incorporating a Laplacian pyramid and a lightweight high-latitude enhancement module, trained jointly to refine feature representations across scales. 
Moreover, the reinforcement learning scheme with latitude-adaptive rewards plays a key role in learning band selection policies, training the network to balance reconstruction fidelity with band pruning. 
However, independent training of each component can cause information disconnection between patches.

\subsection{Diffusion-Based ODISR Methods}

\begin{table*}[!t]
		\rowcolors{1}{gray!20}{white}
		\centering
		\caption{
			{Summary of essential characteristics of representative deep learning-based ODVSR methods. }
		}
		\vspace{-6pt}
		\label{table:odvmethods}
		\begin{threeparttable}
			\resizebox{1\textwidth}{!}{
				\setlength\tabcolsep{2pt}
				\renewcommand\arraystretch{0.98}
                \centering
				\begin{tabular}{c|c||c|c|c|c|c}
					\hline

					&\textbf{Method}&\textbf{Classification} &\textbf{Loss Function}
					&\textbf{Training Data} & \textbf{Testing Data} &\textbf{Evaluation Metric}  \\
					\hline
					\hline
					\multirow{1}{*}{\rotatebox{90}{\textbf{2022}}}
					&VertexShuffle~\cite{VertexShuffle}  &Projection &MSE Loss
					&\tabincell{c}{360-degree Video Dataset} &\tabincell{c}{360-degree Video Dataset} &\tabincell{c}{PSNR\\Parameters Storage Size\\ Per-frame Inference Time} \\
					\hline
					\hline
  
                    \multirow{1}{*}{\rotatebox{90}{\textbf{2023}}}
                    
					&MVideo Team~\cite{2023ntire}  & Traditional VSR &\tabincell{c}{MSE Loss\\Charbonnier \\Loss }&ODV360& ODV360&\tabincell{c}{WS-SSIM\\ WS-PSNR} \\

                    &HIT-IIL~\cite{2023ntire}  & Traditional VSR &Charbonnier loss &ODV360& ODV360&\tabincell{c}{WS-SSIM\\ WS-PSNR} \\
                    
                     &PKU VILLA~\cite{2023ntire}  & Traditional VSR &\tabincell{c}{MSE Loss \\Charbonnier Loss }&ODV360& ODV360&\tabincell{c}{WS-SSIM\\ WS-PSNR} \\                  
					\hline
					\hline
					\multirow{1}{*}{\rotatebox{90}{\textbf{2024}}}
					&STDAN~\cite{STDAN} 
					&Distortion Map &\tabincell{c}{Charbonnier Loss\\ DCT Loss HF Loss  } &ODV-SR Dataset& \tabincell{c}{360VDS \\MiG Panorama Video Dataset} &\tabincell{c}{SSIM PSNR \\WS-SSIM WS-PSNR \\ WE VMAF}\\
                    
    			&SMFN~\cite{SMFN}&Distortion Map
					&\tabincell{c}{Charbonnier Loss \\latitude Loss\\ Saliency Loss}&MiG Panorama Video Dataset& MiG Panorama Dataset&\tabincell{c}{WS-SSIM\\WS-PSNR\\Parameters}\\

					&S3PO~\cite{S3PO} 
					&Distortion Map &WS-L1 Loss &360VDS& \tabincell{c}{360UHD \\360VDS} &\tabincell{c}{SSIM PSNR \\WS-SSIM WS-PSNR \\ Parameters}\\
	
					\hline
					\hline
				\end{tabular}
			}
		\end{threeparttable}
	\end{table*}

The aforementioned studies primarily utilize popular architectures such as CNNs and Transformers to focus on ODISR. These methods largely rely on end-to-end learning strategies, which limit their generalization performance when applied to out-of-distribution data. With the development of diffusion models, image generation methods represented by diffusion models provide strong priors for visual tasks and have been shown to be effective in image restoration tasks. In the ODISR task, diffusion models improve image quality by iteratively denoising from random noise, enabling the reconstruction of fine details and textures in HR images. This process inherently captures rich structural information and facilitates the generation of diverse outputs. 
For example, Li et al.~\cite{omnissr} proposed a zero-shot ODISR method for ODIs by transforming ERP into tangent projection images using stable diffusion priors. In addition, they used iterative refinement with octadecaplex tangent information interaction and gradient decomposition to improve image consistency and quality.
Liu et al.~\cite{diffosr} propose a latent-aware conditional diffusion model to address the challenges of geometric distortion and detail preservation in ODISR. By leveraging latitude distortion maps and high-frequency cues as conditional inputs, the authors effectively model and recover the structural and fine details affected by projection artifacts, leading to significant improvements in SR quality for ODIs.

In summary, these methods leverage the powerful capabilities of deep learning to effectively capture subtle features and contextual information in ODIs, significantly enhancing image clarity and realism. However, these methods also face challenges, such as performance degradation when handling complex scenes and high magnification factors, as well as insufficient adaptability to real-world datasets. Additionally, the consumption of computational resources and the complexity of model training impact deployment in practical applications.

\section{Deep Learning-Based ODVSR}
With the advancements in omnidirectional camera technology, ODVs have increasingly become a prominent medium for recording information. Consequently, many researchers have focused on ODVSR tasks for ODVs to enhance their visual quality. 
Both ODISR and ODVSR aim to enhance the resolution and perceptual quality of wide FOV visual content. Their objective is to recover details and structural information from LR inputs. Both tasks face challenges related to geometric distortions and the preservation of global and local consistency across scenes. However, ODVSR requires the integration of optical flow estimation or temporal information to ensure inter-frame consistency.
In this section, we introduce the ODVSR methods and classification as shown in Table~\ref{table:odvmethods}.

\label{sec:ODVSR}
\subsection{Problem Definition}

Given LR video sequence $V_{LR} = \{ I_{LR}^t \}_{t=1}^{T}$, where $I_{LR}^t\in \mathbb{R}^{W \times H \times 3}$ denotes the LR image with width $W$ and height $H$ in frame $t$, and $T$ is the number of frames. The goal is to generate the SR video sequence $V_{SR} = \{ I_{SR}^t \}_{t=1}^{T}$ from the LR video sequence, where $I_{SR}^t\in \mathbb{R}^{sW \times sH \times 3}$, $s$ denotes the scale factor.  
The ODVSR task can be expressed as:  
\begin{equation}  
V_{SR} = f(V_{LR}, \theta),
\end{equation}  
where $f$ is a mapping function that represents the relationship from LR video to SR video, typically implemented by deep learning models (such as convolutional neural networks, generative adversarial networks, etc.). $\theta$ represents the model parameters, which are optimized through training to minimize the loss between the SR and HR images.

To optimize the above mapping, a loss function $L$ can be applied as follows:  
\begin{equation}  
L = \sum_{t=1}^{T} \mathcal{L}(I_{SR}^t, I_{HR}^t),  
\end{equation}  
where $I_{HR}^t\in \mathbb{R}^{sW \times sH \times 3}$ is the ground truth image at frame $t$. $\mathcal{L}$ denotes the loss function used in the task.

\subsubsection{Distortion Map-Guided ODVSR Methods}  

Current research on ODVSR primarily focuses on addressing the distortions in ERP projections. These methods employ auxiliary information such as distortion maps to effectively mitigate sampling inconsistencies and spatial deformations across different latitudinal regions, thereby enhancing the visual quality and geometric fidelity of the reconstructed videos.

Liu et al.~\cite{SMFN} propose a weighted mean squared error (WMSE) loss function, in which the weights are calculated based on the distortion map. By assigning higher weights to the equatorial areas and lower weights to the polar areas, this approach effectively improves the overall quality of ODVs.
Baniya et al.~\cite{S3PO} propose a weighted spherical smooth L1 loss function to address spherical distortion issues.
These models predominantly focus on latitude-related ERP distortions. They overlook the complex spatiotemporal discontinuities that occur and are perceptually salient. Moreover, most of these methods implicitly assume that viewers are interested in the entire scene uniformly, neglecting the fact that humans typically observe only one local viewpoint at a time.
An et al.~\cite{STDAN} introduce a spatiotemporal distortion modulation module aimed at reducing spatial projection distortions. They also implement a multi-frame reconstruction and fusion mechanism based on internal temporal correlations and alignment strategies to improve frame consistency. Notably, their incorporation of a latitude-saliency adaptive mapping within the loss function prioritized regions with higher texture complexity and human interest, leading to enhanced video restoration quality. 

In summary, these approaches address specific distortion issues in ERP but largely neglect the spatiotemporal dynamics and human viewing behaviors that are crucial for realistic and perceptually aligned ODVS.

\subsubsection{Projection Design-Based ODVSR Methods}  
Building on research focused on distortion map, researchers have also explored innovative projection design methods aimed at optimizing the representation and transmission efficiency of ODVs. These approaches involve designing specialized projection mechanisms to maintain high-quality spherical content while considering bandwidth and computational resource constraints.
Li et al.~\cite{VertexShuffle} design a new Spherical Super-Resolution (SSR) method. The method utilizes a focused icosahedral mesh to represent a small area on the sphere and constructs a matrix to rotate spherical content into the focused mesh region, in order to address bandwidth waste and computational load associated with ODVs transmission.

\begin{table}[t]  
    \centering  
    \caption{Summary of ODISR and ODVSR datasets.}
    \renewcommand{\arraystretch}{1.5} 
    \fontsize{8.5pt}{9.5pt}\selectfont  
    \tabcolsep=3.5pt  
    \begin{tabular}{c|ccc|c}  
        \hline  
        \textbf{Name} & \textbf{\#Train} & \textbf{\#Val} & \textbf{\#Test} & \textbf{Resolution} \\
        \hline  
        \multicolumn{5}{c}{\textbf{ODI Datasets}} \\
        \hline  
        ODISR \cite{lau}     & 1200 & 100 & 100 & 2048$\times$1024 \\
        \hline  
        ODISR-clean \cite{OSRT}     & 1150 & 97 & 100 & 2048$\times$1024 \\
        \hline  
        SUN360 \cite{lau}      & -    & -   & 100 & 2048$\times$1024 \\
        \hline  
        SUN360-clean \cite{OSRT}      & -    & -   & 100 & 2048$\times$1024 \\
        \hline  
        Flickr360 \cite{2023ntire}  & 3000 & 50  & 100 & 2048$\times$1024 \\
        \hline  
        360Insta (OURS)  & - & -  & 1500 & 3840$\times$1920 \\
        \hline  
        \multicolumn{5}{c}{\textbf{ODV Datasets}} \\
        \hline  
        MiG Panorama \cite{SMFN}& 200  & -   & 8   & 1440$\times$720$-$4096$\times$2048 \\
        \hline  
        360VDS \cite{S3PO}     & 545  & -  & 45   & 480$\times$360 \\
        \hline  
        360HUD \cite{S3PO}      & -    & -   & 8   & 1280$\times$720$-$4096$\times$2048 \\
        \hline  
        ODV-SR \cite{STDAN}      & 270    &20    & 25   & 2160$\times$1080 \\
        \hline  
        ODV360 \cite{2023ntire}     & 210  & 20  & 20  & 2160$\times$1080 \\
        \hline  
    \end{tabular}  
  
    \label{tab:datasets}  
\end{table}

\subsubsection{Traditional VSR-Based ODVSR Methods} 
In addition to the ODVSR frameworks mentioned above, the 2023 NTIRE challenge competition~\cite{2023ntire} featured three notable ODVSR methods as follows: MVideo Team, HIT-IIL Team, and PKU VILLA Team. 
MVideo Team utilized the BasicVSR++ framework~\cite{basicvsr++}, which employs flow-guided deformable alignment to reduce the burden of offset learning by using optical flow fields as the basis for offsets. 
HIT-IIL Team similarly recognized that ERP videos stretch the regions near the poles horizontally. They adopted BasicVSR++~\cite{basicvsr++} as the perspective VSR network, employing second-order grid propagation for improved temporal modeling and flow-guided deformable alignment for stable offset estimation. 
PKU VILLA Team enhanced the model's SR capabilities through a two-stage fusion strategy and parameter tuning.

In summary, current research on ODVSR is still relatively limited, mainly focusing on traditional VSR models, distortion map, and exploration of spherical projection optimization methods. Although these technologies have made some progress, there is still a lot of room for improvement, and it is expected that deeper breakthroughs will be achieved in data, models, and algorithms.

\section{Datasets}
\label{sec:dataset}

Table~\ref{tab:datasets} presents the datasets currently used for ODIs and ODVs. 
These datasets exhibit diverse characteristics in terms of content richness and resolution, serving as essential resources for training, validation, and performance evaluation of algorithms. 
Most of the datasets are based on HR acquisitions, with LR data generated via simple bicubic downsampling. While convenient, this approach often fails to accurately simulate real-world degradations, thus limiting the evaluation of model robustness under more realistic conditions. To address this gap, we propose a novel dataset that incorporates more practical and complex degradation processes, providing a more reliable benchmark for assessing model robustness.
The following section introduces several representative datasets and our proposed 360Insta dataset, including ODISR~\cite{lau}, ODISR-clean~\cite{OSRT}, SUN360~\cite{lau}, SUN360-clean~\cite{OSRT}, and Flickr360~\cite{2023ntire}, as well as ODV datasets such as MiG Panorama~\cite{SMFN}, 360VDS~\cite{S3PO}, ODV-SR~\cite{STDAN}, and ODV360~\cite{2023ntire}, highlighting their unique features and addressing different needs within the scope of ODISR and ODVSR.


\subsection{ODI Datasets}  

\begin{figure*}[!t]  
    \centering  
    \includegraphics[width=0.98\textwidth]{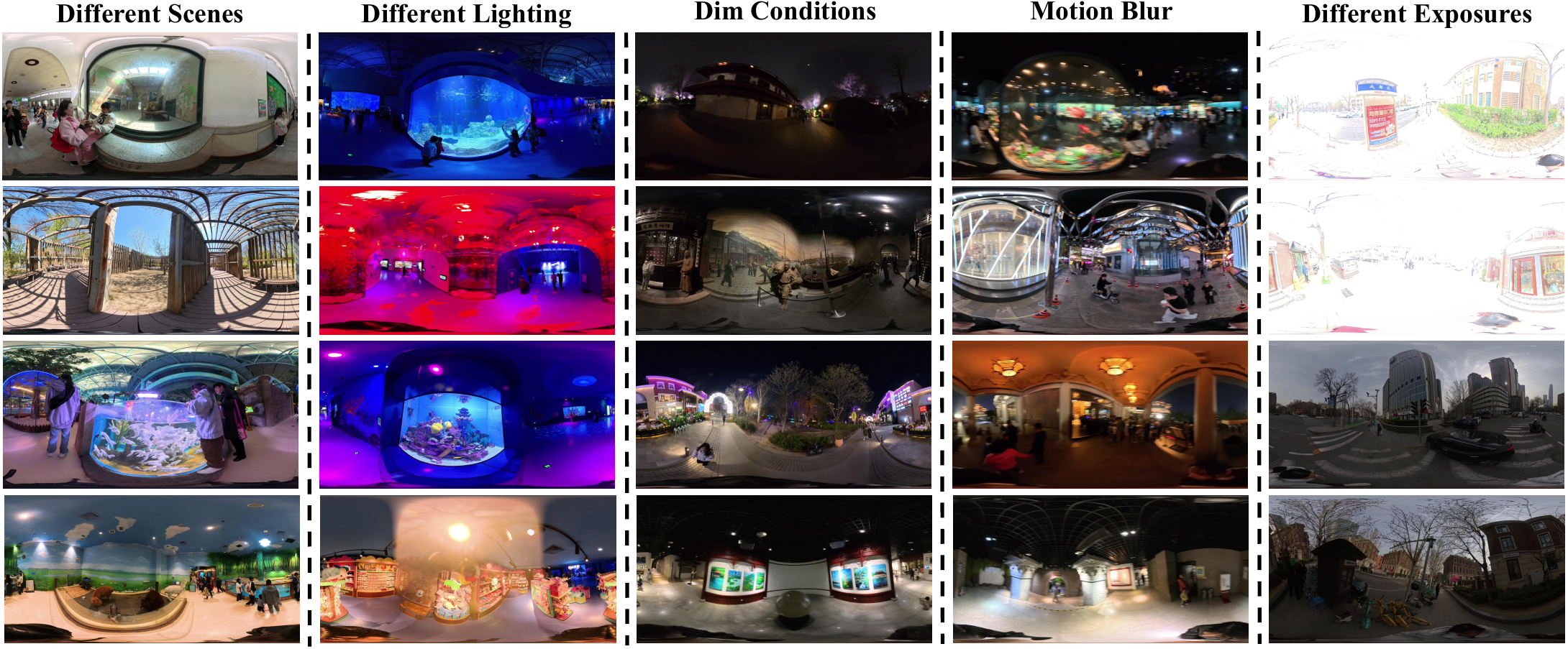}  
    \caption{Illustrations of the ODIs with different image content in our 360Insta dataset. 360Insta includes different scenes (e.g., nature landscape, indoor landscape, and exhibition scenarios), different lighting, dim conditions, motion blur, and different exposures.}  
    \label{fig:360insta}  
\end{figure*} 

\textbf{ODISR Dataset\footnote{\url{https://github.com/wangh-allen/LAU-Net}}}: The ODISR dataset~\cite{lau} comprises 1,400 HR ODIs, including ODIs collected from FlickrVR~\cite{vrdataset} and the Internet, with a resolution of 2048$\times$1024. LR ODIs are generated by bicubic downsampling with a specific downsampling factor. For image content, ODIs are chosen to cover a variety of visual themes such as natural scenes, human activities, exhibitions, etc. The number of ODIs for urban landscapes, natural landscapes, indoor scenes, human activities, and exhibitions is 245, 301, 310, 294, and 250, respectively. The image content encompasses the most common ODI photographic scenes. 
When dividing the dataset, an even distribution of images with different content types is maintained among the training, validation, and testing sets, which include 1,200, 100, and 100 images, respectively.
Based on the ODISR dataset, OSRT method~\cite{OSRT} proposes to apply fisheye downsampling by an anti-aliased bicubic function and store images in a lossless format (PNG). Moreover, they observed problems in previous datasets: 1) transforming mistakes; 2) virtual scenarios; 3) extremely low qualities; 4) plane images. They cleaned the data, that is, they selected the problematic data sets mentioned above. Consequently, they propose ODISR-clean~\cite{OSRT} datasets, consisting of 1,150 images for training, 97 images for validation, and 100 images for testing.

\noindent
\textbf{SUN360 Dataset}: The SUN360 dataset~\cite{sun360} contains 80 categories and 67,583 ODIs, using ERP to cover 360°$\times$180° viewing angles. 
For the ODISR task, the LAU-Net method~\cite{lau} extracts 100 ODIs for testing with 2048°$\times$1024. Based on the LAU-Net method~\cite{lau}, OSRT~\cite{OSRT} designs the SUN360-clean dataset applying fisheye downsampling by an anti-aliased bicubic function.

\noindent
\textbf{Flickr360 Dataset\footnote{\url{https://github.com/360SR/360SR-Challenge}}}: The Flickr360 dataset contains 3,150 ERP images, all with an original resolution greater than 5k. Specifically, 3,100 images were collected from Flickr \footnote{\url{https://www.flickr.com/}}, and the remaining 50 images were captured using an Insta360° omnidirectional camera. The image content varies between indoor and outdoor environments, featuring many natural scenes, human constructions, and street views. In terms of data pairing, the original images are first resized to 2k resolution (2048$\times$1024) as HR images. These HR images are then further downsampled to produce LR images. The images are randomly divided into training, validation, and testing sets, which include 3,000 images, 50 images, and 100 images, respectively.

\subsection{ODV Datasets}  
\textbf{MiG Panorama Dataset\footnote{\url{https://drive.google.com/drive/folders/1CcBiblzkHVXZ1aSdSdZgvGPUkWuoMsE1}}}: The MiG Panorama dataset~\cite{SMFN} includes 208 ODVs in ERP. Each video contains 100 frames with resolutions ranging from 4096$\times$2048 to 1440$\times$720. The scenes include both outdoor and indoor environments. Eight representative videos were selected and divided into two sets for testing, the rest being used for training. 
Due to limited computational resources, each video was initially downscaled by a factor of 2, resulting in resolutions between 2048$\times$1024 to 720$\times$360, using bicubic interpolation to generate the HR ground truth. Subsequently, the ground truth videos are further downsampled by a factor of 4 with the same interpolation method to produce the LR counterparts.

\noindent
\textbf{360VDS Dataset}: The 360VDS dataset~\cite{S3PO} is assembled from open-source datasets used in other ODV research, such as quality assessment~\cite{bridge}, compression~\cite{learning}, saliency modeling~\cite{saliency}, and literature reviews~\cite{survey,sota,taxonomy}. Additionally, it uses publicly available ODV datasets from the Stanford VR Lab, known as psych-360~\cite{personal}. The 360VDS dataset comprises a total of 590 20-frame ODVs, all resized to a uniform resolution of 480$\times$360 pixels to address size constraints during training of the recurrent model. The 360VDS dataset is randomly split into 45 videos for the test set and 545 videos for the training set.

\noindent
\textbf{360UHD Dataset}: The 360UHD dataset~\cite{S3PO} consists of eight clips, which are HR videos used for testing. These clips are randomly selected from the 45 clips of the 360VDS test set, with the original HR ground truth values kept unchanged, ranging from 1280$\times$720$-$4096$\times$2048.

\noindent
\textbf{ODV-SR Dataset}: 
The ODV-SR dataset~\cite{STDAN} contains 70 high-quality ODVs collected from YouTube and public 360° video datasets, which have a resolution of 2160$\times$1080, collecting not only relatively static scenes such as landscapes, concerts, and narratives, but also dynamic scenes such as sports, driving, and movies. 
In addition to these diverse categories, the ODV-SR dataset incorporates ODVs derived from gaming content, animations, and other virtual environments. To enhance the diversity and comprehensiveness of the dataset, these samples are combined with the ODV360 dataset~\cite{2023ntire}, thereby extending its variability and coverage. The ODV-SR data set is then divided into training, validation, and testing sets, which include 270 clips, 20 clips, and 25 clips, respectively. The LR ODVs of these videos are generated using bicubic downsampling.

\noindent
\textbf{ODV360 Dataset\footnote{\url{https://drive.google.com/drive/folders/1lDIxTahDXQ5w5x_UZySX2NOes_ZoNztN}}}: The ODV360 dataset~\cite{2023ntire} consists of 90 HR videos collected from YouTube and public ODV datasets, intended for restoration. Moreover, 160 videos were collected using Insta360 cameras, including the Insta 360$\times$2 and Insta 360 ONE RS, which capture HR (5720$\times$3210) omnidirectional videos. 
The ODVs collected demonstrate substantial diversity, covering a range of indoor and outdoor scenes. To ensure uniformity, all videos were downsampled to a resolution of 2K (2160$\times$1080), with each clip consisting of 100 frames. 
These videos are randomly divided into training, validation, and testing sets, comprising 210 clips, 20 clips, and 20 clips.


\subsection{Proposed 360Insta Dataset}
Most existing omnidirectional datasets synthesize LR images by downsampling HR ground truths. However, this synthetic approach does not accurately reflect real-world degradation processes, thus limiting the assessment of a model’s robustness. 
To address this gap, we propose the 360Insta dataset, captured using the Insta360 $\times$4 camera, which features a sensor resolution of 3840$\times$1920 and a video sampling rate of 24 fps. The camera is equipped with a 360°$\times$180° omnidirectional wide-angle lens. The dataset encompasses a wide variety of scenes and content, covering different environmental conditions such as indoor, outdoor, urban, and natural landscapes. The recordings span both daytime and nighttime scenarios, showcasing the dataset’s rich diversity and representativeness.

To enrich scene complexity, data are collected using a range of techniques, including handheld selfie sticks, stationary tripods, and dynamic mobile captures. For example, scene blurring degradation was simulated through handheld and jittery movements, while exposure levels were varied by adjusting the exposure compensation parameter to produce images under different lighting conditions. All footage is processed using Insta360 Studio and exported in the form of H.264-encoded ODIs and ODVs. To realistically reflect common degradation phenomena, the dataset categorizes various types of degradations, including different lighting conditions, dim conditions, motion blur, and different exposure settings.

Specifically, as shown in Fig.~\ref{fig:360insta}, the dataset includes: 698 images depicting multi-scene blurring, 260 images under different lighting conditions, 283 images captured in dim conditions, 141 images with motion blur, and 118 images with different exposure settings. These authentic and diverse degraded scenes provide a significant advantage in simulating real-world scenarios, offering a more real benchmark for model robustness evaluation. Compared to synthetic datasets, 360Insta offers higher realism and diversity, furnishing a more reliable platform for research in ODISR.
The dataset has been publicly released at \url{https://github.com/nqian1/Survey-on-ODISR-and-ODVSR} to facilitate further research and development in ODISR and ODVSR and related applications.

\begin{table*}[!t]\scriptsize
\center
\begin{center}
\caption{Comparison of Deep Learning-based ODISR Methods under ERP Downsampling on ODISR and SUN360 Datasets~\cite{lau}. $\uparrow$: the higher, the better. The best result is in {\color{red}{red}} whereas the second and third best results are in {\color{blue}{blue}} and {\color{green}{green}} under each case, respectively.}

\label{tab:ori_odisr_results}
\small
\footnotesize
\fontsize{8.5pt}{9.5pt}\selectfont
\tabcolsep=6pt
\begin{tabular}{c|c|cccc|cccc}
\hline
\multirow{2}{*}{Method} & \multirow{2}{*}{Scale} & \multicolumn{4}{c|}{ODISR~\cite{lau}} & \multicolumn{4}{c}{SUN360~\cite{sun360}}
\\
\cline{3-10}
& &  PSNR $\uparrow$ & SSIM $\uparrow$& WS-PSNR $\uparrow$& WS-SSIM $\uparrow$& PSNR $\uparrow$& SSIM $\uparrow$& WS-PSNR $\uparrow$& WS-SSIM $\uparrow$\\
\hline

360-SS\cite{360SS} & $\times$2 
& {\color{red}27.89}
& {\color{red}0.8240}
& 24.17
& {\color{blue}0.8094}
& {\color{red}28.48}
& {\color{red}0.8274}
& {\color{red}28.02}
& {\color{red}0.8292}
\\

OSRT-light~\cite{OSRT} \quad\quad & $\times$2 
& {\color{green}25.89}
& {\color{blue}0.7901}
& {\color{green}26.30}
& {\color{green}0.8106}
& {\color{green}25.49}
& {\color{green}0.7704}
& {\color{green}26.63}
& {\color{blue}0.8238}
\\
OSRT~\cite{OSRT} & $\times$2 
& 25.08
& {\color{green}0.7740}
& {\color{blue}25.71}
& {\color{red}0.8002}
& 24.34
& 0.7438
& 25.82
& {\color{green}0.8092}
\\

OmniSSR~\cite{omnissr} & $\times$2
& {\color{blue}26.99}
& 0.7731
& {\color{red}26.36}
& 0.7655
& {\color{blue}27.33}
& {\color{blue}0.7756}
& {\color{blue}27.24}
& 0.7986
\\
\hline
360-SS\cite{360SS} & $\times$4 
& {\color{green}23.95}
& {\color{green}0.6873}
& {\color{green}23.21}
& 0.6604
& {\color{green}23.84}
& {\color{green}0.6732}
& {\color{green}23.21}
& 0.6710
\\

OSRT-light~\cite{OSRT} & $\times$4 
& 22.34
& 0.6665
& 22.60
& {\color{green}0.6735}
& 21.06
& 0.6436
& 22.35
& {\color{green}0.6874}
\\
OSRT~\cite{OSRT} & $\times$4 
& 21.75
& 0.6467
& 22.23
& 0.6617
& 20.94
& 0.6083
& 21.92
& 0.6653
\\

{BPOSR}\cite{BPOSR} & $\times$4 
& {\color{red}26.35}
& {\color{red}0.7551}
& {\color{red}25.53}
& {\color{red}0.7312}
& {\color{red}26.79}
& {\color{red}0.7670}
& {\color{red}26.10}
& {\color{red}0.7725}
\\
OmniSSR~\cite{omnissr} & $\times$4
& {\color{blue}25.59}
& {\color{blue}0.7222}
& {\color{blue}24.87}
& {\color{blue}0.7047}
& {\color{blue}25.52}
& {\color{blue}0.7163}
& {\color{blue}25.07}
& {\color{blue}0.7279}

\\ \hline
360-SS\cite{360SS} & $\times$8 
& 21.30
& 0.6089
& 20.53
& 0.5708
& 20.94
& 0.5878
& 20.19
& 0.5764
\\
LAU\cite{lau}& $\times$8 
& {\color{green}23.53}
& {\color{blue}0.6556}
& {\color{green}22.90}
& {\color{blue}0.6309}
& {\color{green}23.33}
& {\color{blue}0.6563}
& {\color{green}22.71}
& {\color{blue}0.6588}
\\
{BPOSR}\cite{BPOSR} & $\times$8 
& {\color{red}24.01}
& {\color{red}0.6730}
& {\color{blue}23.18}
& {\color{red}0.6409}
& {\color{red}23.84}
& {\color{red}0.6775}
& {\color{blue}22.99}
& {\color{red}0.6703}
\\
{OmniSSR}\cite{omnissr} & $\times$8
& {\color{blue}23.99}
& {\color{green}0.6502}
& {\color{red}23.26}
& {\color{green}0.6209}
& {\color{blue}23.55}
& {\color{green}0.6384}
& {\color{red}23.00}
& {\color{green}0.6356}
\\ \hline
{BPOSR}\cite{BPOSR} & $\times$16 
& {\color{blue}22.05}
& {\color{red}0.6258}
& {\color{blue}21.21}
& {\color{red}0.5881}
& {\color{blue}21.49}
& {\color{red}0.6159}
& {\color{blue}20.61}
& {\color{red}0.6020}
\\
{OmniSSR}\cite{omnissr} & $\times$16 
& {\color{red}22.56}
& {\color{blue}0.6119}
& {\color{red}21.82}
&{\color{blue} 0.5751}
& {\color{red}21.94}
& {\color{blue}0.5979}
& {\color{red}21.32}
& {\color{blue}0.5851}
\\ \hline
\end{tabular}
\end{center}
\end{table*}

\begin{table*}[ht]\scriptsize
\center
\begin{center}
\caption{Comparison of Deep Learning-based ODISR Methods under Fisheye Downsampling on ODISR-clean and SUN360-clean Datasets~\cite{OSRT}. $\uparrow$: the higher, the better. The best result is in {\color{red}{red}} whereas the second and third best results are in {\color{blue}{blue}} and {\color{green}{green}} under each case, respectively.}

\label{tab:fish_odisr_results}
\small
\footnotesize
\fontsize{8.5pt}{9.5pt}\selectfont
\tabcolsep=6pt
\begin{tabular}{c|c|cccc|cccc}
\hline
\multirow{2}{*}{Method} & \multirow{2}{*}{Scale}  & \multicolumn{4}{c|}{ODISR-clean~\cite{lau}} & \multicolumn{4}{c}{SUN360-clean~\cite{sun360}}
\\
\cline{3-10}
& & PSNR $\uparrow$& SSIM $\uparrow$& WS-PSNR $\uparrow$& WS-SSIM $\uparrow$& PSNR $\uparrow$& SSIM $\uparrow$& WS-PSNR $\uparrow$& WS-SSIM $\uparrow$\\
\hline
360-SS\cite{360SS}& $\times$2 
& {\color{green}28.85}
& {\color{green}0.8423}
& {\color{green}28.31}
& {\color{green}0.8407}
& {\color{green}29.05}
& {\color{green}0.8379}
& {\color{green}29.20}
& {\color{green}0.8606}
\\
OSRT-light~\cite{OSRT} \quad\quad & $\times$2 
&  {\color{blue}30.36}
&  {\color{blue}0.8766}
&  {\color{blue}29.72}
&  {\color{blue}0.8723}
&  {\color{blue}30.92}
&  {\color{blue}0.8781}
&  {\color{blue}31.44}
&  {\color{blue}0.8989}
\\
OSRT~\cite{OSRT} & $\times$2 
& {\color{red}30.70}
& {\color{red}0.8836}
& {\color{red}30.03}
& {\color{red}0.8781}
& {\color{red}32.02}
& {\color{red}0.8878}
& {\color{red}31.64}
& {\color{red}0.9066}
\\
OmniSSR~\cite{omnissr} & $\times$2
& 26.46
& 0.7549
& 25.87
& 0.7478
& 26.81
& 0.7554
& 26.72
& 0.7798
\\

\hline
360-SS\cite{360SS}& $\times$4 
& 25.96
& 0.7246
& 25.32
& 0.7091
& 25.79
& 0.7167
& 25.53
& 0.7298
\\

OSRT-light~\cite{OSRT} & $\times$4 
& {\color{blue}27.15}
& {\color{blue}0.7664}
& {\color{blue}26.47}
& {\color{blue}0.7522}
& {\color{blue}27.46}
& {\color{blue}0.7713}
& {\color{blue}27.37}
& {\color{green}0.7904}
\\
OSRT~\cite{OSRT} & $\times$4 
& {\color{red}27.37}
& {\color{red}0.7756}
& {\color{red}26.66}
& {\color{red}0.7602}
& {\color{red}27.80}
& {\color{red}0.7829}
& {\color{red}27.73}
& {\color{red}0.8011}
\\

{BPOSR}\cite{BPOSR} & $\times$4 
& {\color{green}26.88}
& {\color{green}0.7576}
& {\color{green}26.22}
& {\color{green}0.7440}
& {\color{green}27.02}
& {\color{green}0.7594}
& {\color{green}27.36}
& {\color{blue}0.7916}
\\ 
OmniSSR~\cite{omnissr} & $\times$4
& 23.76
& 0.6820
& 23.00
& 0.6605
& 23.69
& 0.6718
& 23.10
& 0.6785
\\
\hline
360-SS\cite{360SS}& $\times$8 
& {\color{red}24.27}
& {\color{red}0.6560}
& {\color{red}23.58}
& {\color{red}0.6279}
& {\color{red}23.93}
& {\color{blue}0.6460}
& {\color{red}23.47}
& {\color{blue}0.6450}
\\
LAU\cite{lau} & $\times$8 
& {\color{green}23.83}
& {\color{green}0.6409}
& {\color{green}23.30}
& {\color{green}0.6162}
& {\color{green}23.33}
& {\color{red}0.6563}
& {\color{green}22.71}
& {\color{red}0.6588}
\\
{BPOSR}\cite{BPOSR} & $\times$8 
& {\color{blue}24.15}
& {\color{blue}0.6524}
& {\color{blue}23.42}
& {\color{blue}0.6213}
& {\color{blue}23.67}
& {\color{green}0.6430}
& {\color{blue}23.15}
& {\color{green}0.6403}
\\ 
OmniSSR~\cite{omnissr} & $\times$8
& 21.87
& 0.6177
& 21.14
& 0.5866
& 21.48
& 0.6029
&  20.81
& 0.5968
\\
\hline
{BPOSR}\cite{BPOSR} & $\times$16
& {\color{red}22.54}
& {\color{red}0.6168}
& {\color{red}21.77}
& {\color{red}0.5791}
& {\color{red}21.94}
& {\color{red}0.6050}
& {\color{red}21.30}
& {\color{red}0.5934}
\\ 
OmniSSR~\cite{omnissr} & $\times$16
& {\color{blue}20.37}
& {\color{blue}0.5875}
& {\color{blue}19.63}
& {\color{blue}0.5500}
& {\color{blue}19.74}
& {\color{blue}0.5694}
& {\color{blue}19.02}
& {\color{blue}0.5543}
\\
\hline
\end{tabular}
\end{center}
\end{table*}

\begin{table}[ht]  
\centering  
\caption{Comparison of parameters across various models on the ODISR-clean test set with a $\times$4 upscaling factor. FLOPs (G) are calculated based on the LR input with a resolution of 512$\times$256. }
\label{tab:param}  
\setlength{\tabcolsep}{10pt} 
\begin{tabular}{c|c|c}  
\hline  
Method & Params (M) & FLOPs (G) \\
\hline  
OSRT~\cite{OSRT} & 14.2 & 115 \\
BPOSR~\cite{BPOSR} & 2.07 & 32 \\
360-SS~\cite{360SS} & 1.6 & 14.8 \\
SwinIR~\cite{swinsr} & 12.4 & 94.6 \\
LAU-Net~\cite{lau} & 9.4 & 102.8 \\
\hline  
\end{tabular}  
\end{table}

\begin{table*}[ht]\scriptsize
\center
\begin{center}
\caption{Comparison of Deep Learning-based ODVSR Methods under Fisheye Downsampling on MiG Panorama~\cite{SMFN} and 360VDS~\cite{S3PO} Datasets. $\uparrow$: the higher, the better. The best result is in {\color{red}{red}} whereas the second and third best results are in {\color{blue}{blue}} and {\color{green}{green}} under each case, respectively.}

\label{tab:fish_odvsr_results}
\small
\footnotesize
\fontsize{8.5pt}{9.5pt}\selectfont
\tabcolsep=6pt
\begin{tabular}{c|c|cccc|cccc}
\hline
\multirow{2}{*}{Method} & \multirow{2}{*}{Scale} & \multicolumn{4}{c|}{ MiG Panorama \cite{VertexShuffle}} & \multicolumn{4}{c}{360VDS \cite{S3PO}}
\\
\cline{3-10}
&  & PSNR $\uparrow$& SSIM $\uparrow$& WS-PSNR $\uparrow$& WS-SSIM $\uparrow$& PSNR $\uparrow$& SSIM $\uparrow$& WS-PSNR $\uparrow$& WS-SSIM $\uparrow$\\
\hline

Aalign~\cite{align} & $\times$4 
& {\color{blue}28.55}
& {\color{red}0.8836}
& {\color{blue}28.07}
& {\color{blue}0.7983}
& {\color{red}25.22}
& {\color{blue}0.7945}
& {\color{blue}24.46}
& {\color{blue}0.7663}
\\

IART\cite{IART}& $\times$4 
& 28.17
& 0.8121
& 27.59
& 0.7861
& {\color{blue}25.16}
& {\color{green}0.7929}
& {\color{green}24.40}
& {\color{green}0.7643}
\\

MIAVSR$^\dagger$~\cite{MIAVSR} & $\times$4 
& {\color{green}28.54}
& {\color{green}0.8219}
& {\color{red}28.08}
& {\color{green}0.7972}
& {\color{red}25.22}
& {\color{blue}0.7949}
& {\color{red}24.47}
& {\color{red}0.7669}
\\
RealBasicVSR~\cite{RealBasicVSR} & $\times$4 
& 26.62
& 0.7760
& 26.29
& 0.7445
& 23.32
& 0.7121
& 22.58
& 0.6775
\\

{SAVSR}\cite{SAVSR} & $\times$4 
& {\color{red}29.68}
& {\color{blue}0.8447}
& 29.12
& {\color{red}0.8227}
& 25.05
& 0.7837
& 24.25
& 0.7531

\\ \hline
\end{tabular}
\end{center}
\end{table*}

\begin{figure*}[!t]  
    \centering  
    \includegraphics[width=0.98\textwidth]{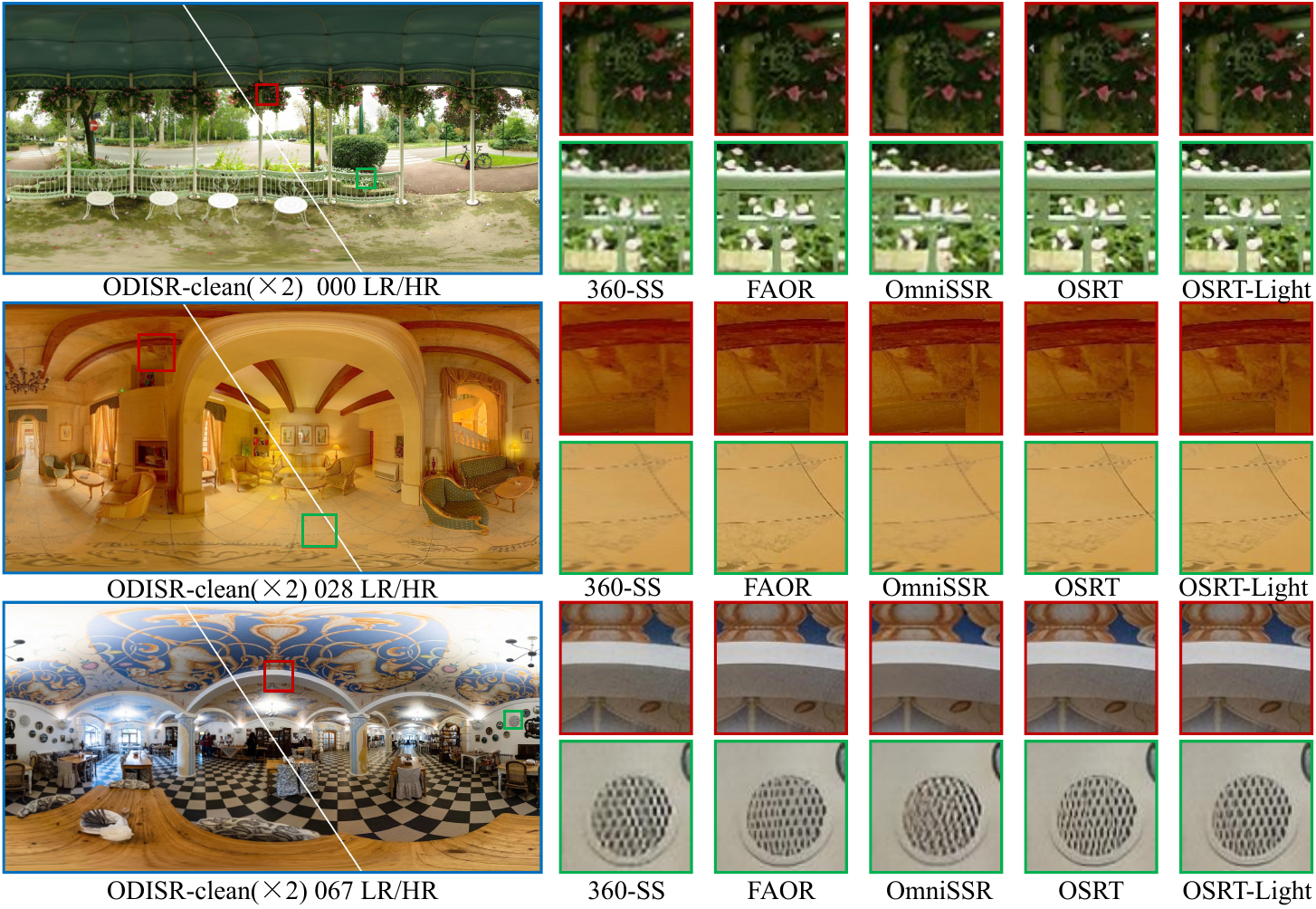}  
    \caption{Visual comparison of the super-resolution results of 360-SS~\cite{360SS}, FAOR~\cite{FAOR}, OmniSSR~\cite{omnissr}, OSRT~\cite{OSRT}, and OSRT-light~\cite{OSRT} on the ODISR-clean~\cite{OSRT} dataset at $\times$2.}  
    \label{fig:x2odisr}  
\end{figure*}  

\begin{figure*}[h]  
    \centering  
    \includegraphics[width=0.98\textwidth]{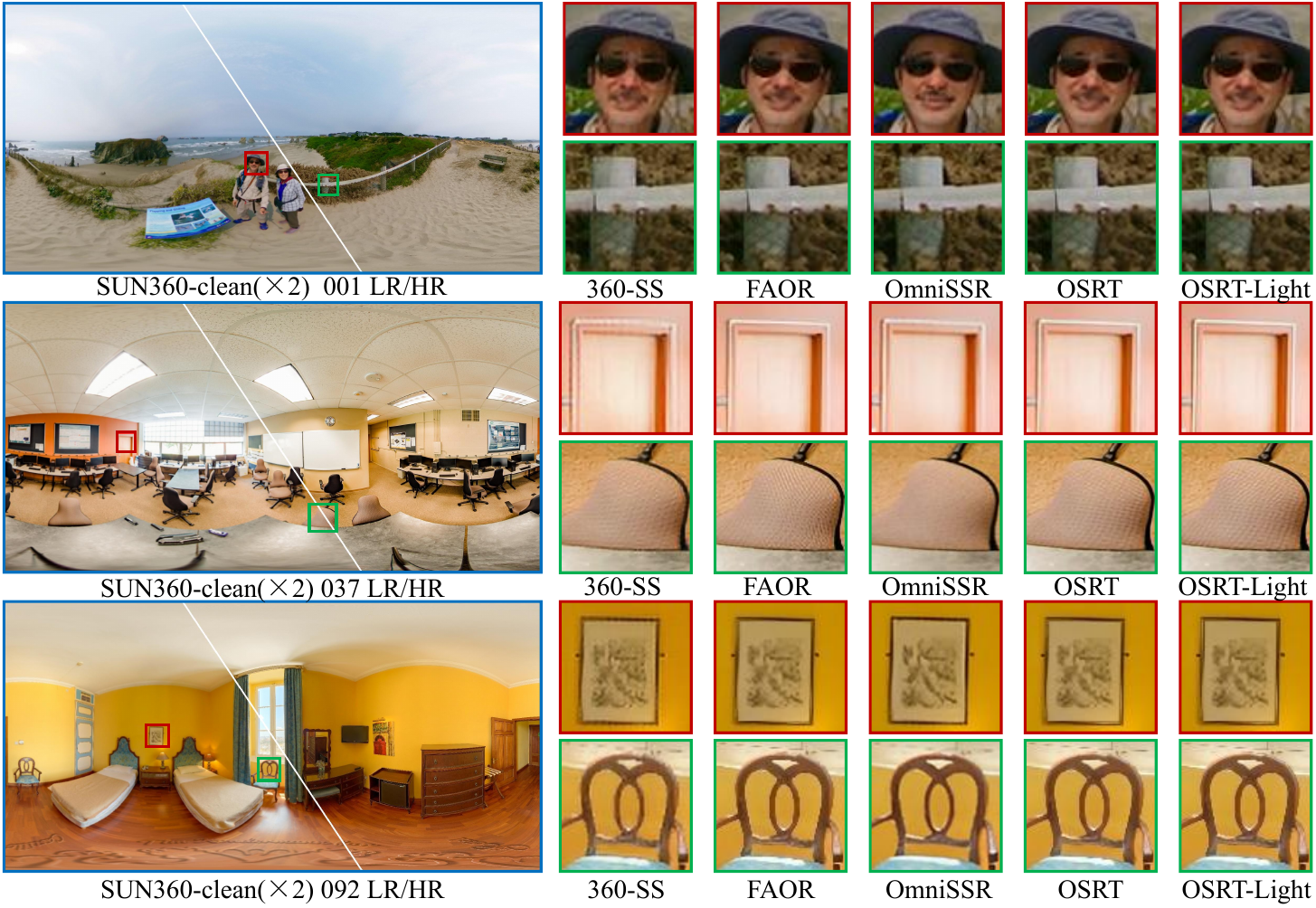}  
    \caption{Visual comparison of the super-resolution results of 360-SS~\cite{360SS}, FAOR~\cite{FAOR}, OmniSSR~\cite{omnissr}, OSRT~\cite{OSRT}, and OSRT-light~\cite{OSRT} on the SUN360-clean~\cite{OSRT} dataset at $\times$2.}  
    \label{fig:x2sun}  
\end{figure*}  

\begin{figure*}[ht]  
    \centering  
    \includegraphics[width=0.98\textwidth]{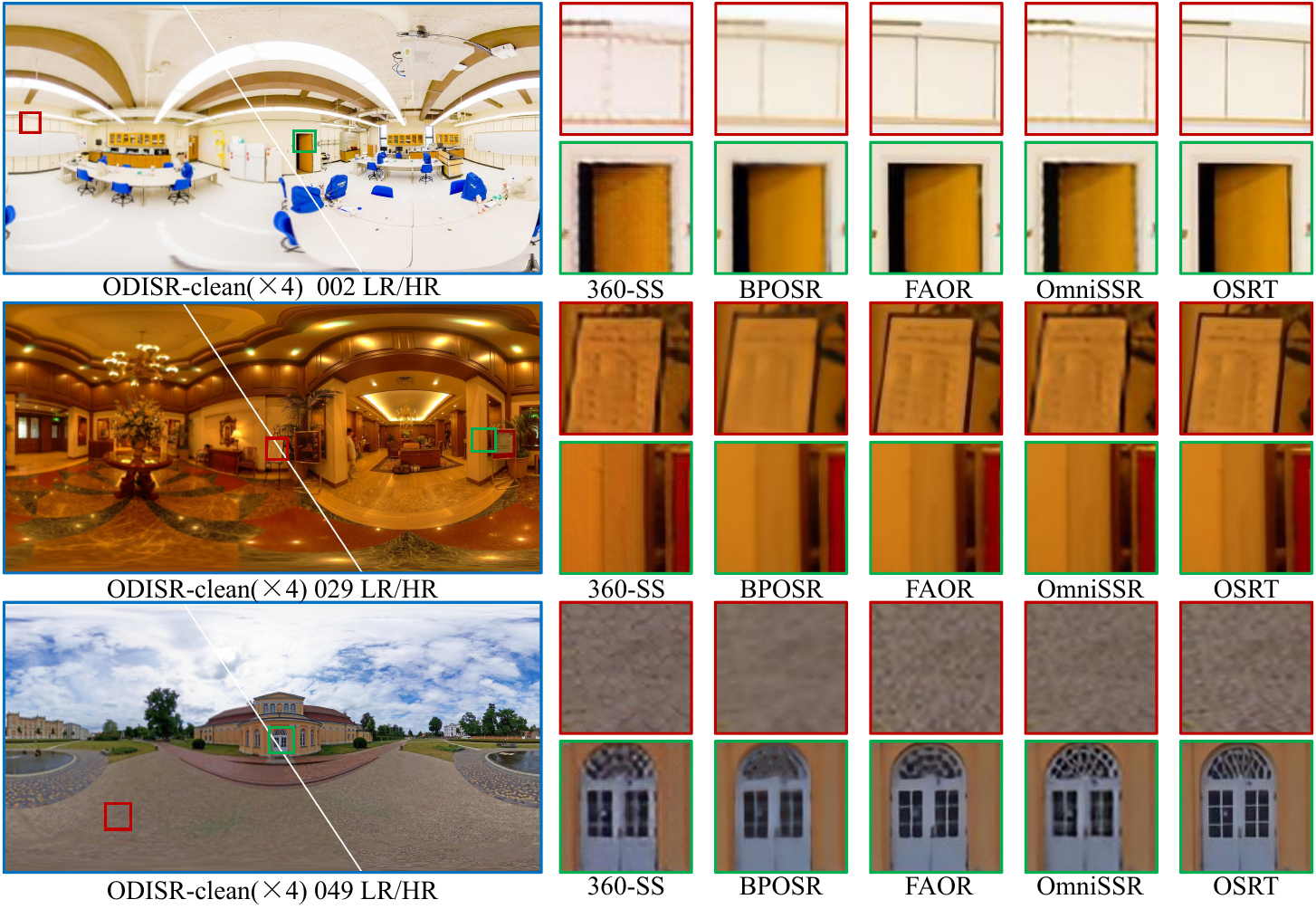}  
    \caption{Visual comparison of the super-resolution results of 360-SS~\cite{360SS}, BPOSR~\cite{BPOSR}, FAOR~\cite{FAOR}, OmniSSR~\cite{omnissr}, and OSRT~\cite{OSRT} on the ODISR-clean~\cite{OSRT} dataset at $\times$4.}   
    \label{fig:x4odisr}  
\end{figure*}

\begin{figure*}[ht]  
    \centering  
    \includegraphics[width=0.98\textwidth]{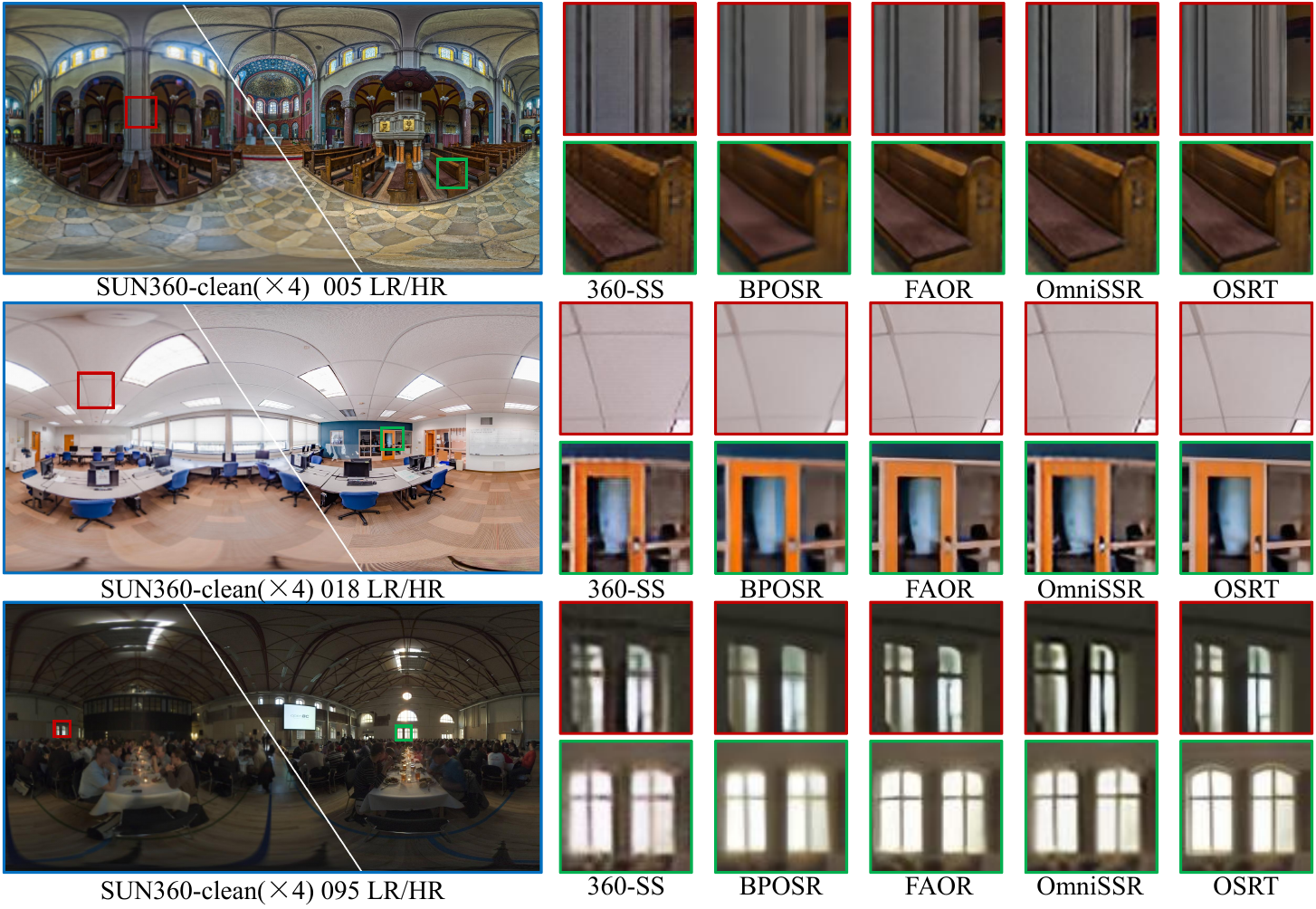}  
    \caption{Visual comparison of the super-resolution results of 360-SS~\cite{360SS}, BPOSR~\cite{BPOSR}, FAOR~\cite{FAOR}, OmniSSR~\cite{omnissr}, and OSRT~\cite{OSRT} on the SUN360-clean~\cite{OSRT} dataset at $\times$4.}  
    \label{fig:x4sun}  
\end{figure*}

\begin{figure*}[t]  
    \centering  
    \includegraphics[width=0.98\textwidth]{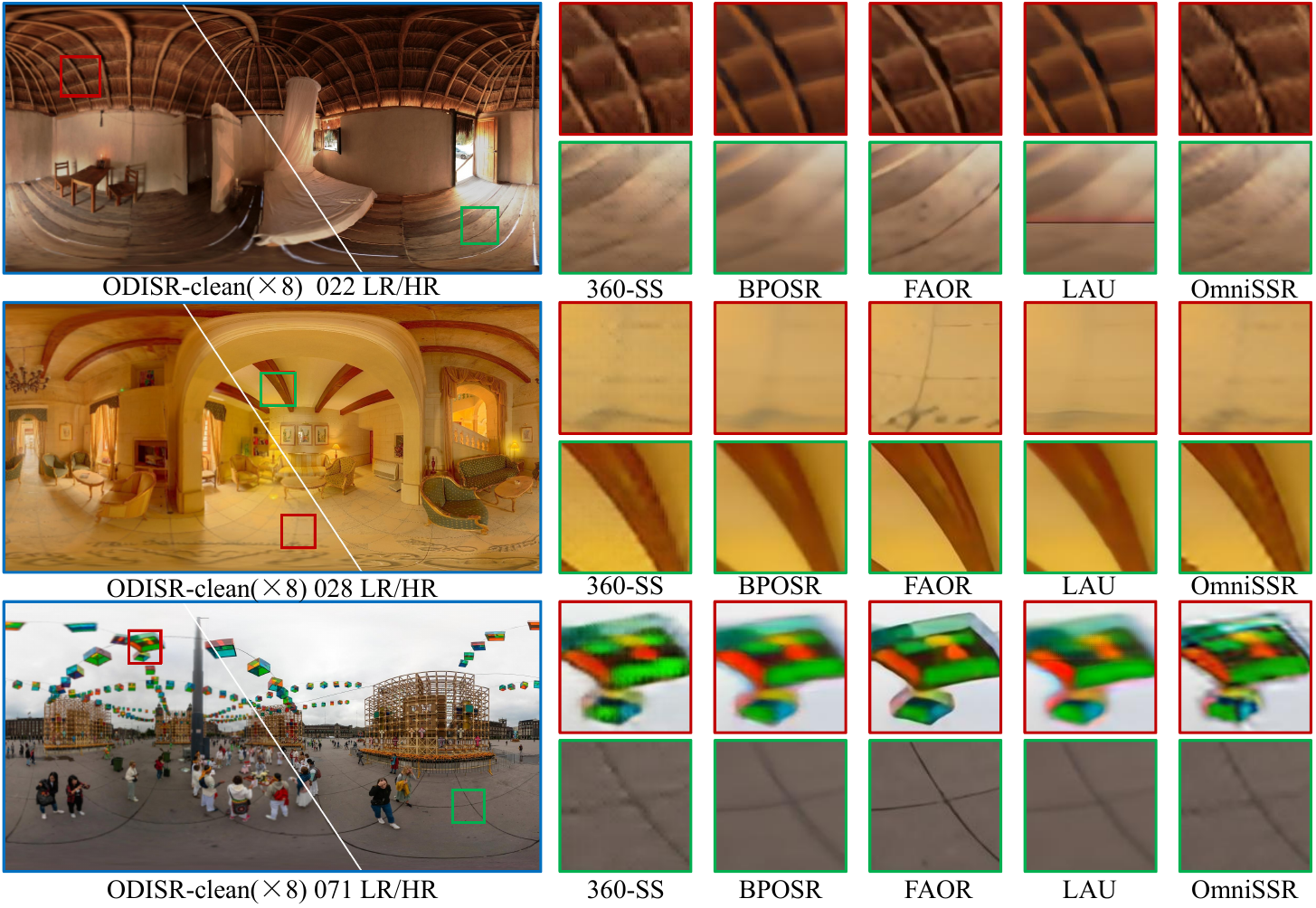}  
    \caption{Visual comparison of the super-resolution results of 360-SS~\cite{360SS}, BPOSR~\cite{BPOSR}, FAOR~\cite{FAOR}, LAU~\cite{lau}, and OmniSSR~\cite{omnissr} on the ODISR-clean~\cite{OSRT} dataset at $\times$8.}  
    \label{fig:x8odisr}  
\end{figure*}  

\begin{figure*}[!t]  
    \centering  
    \includegraphics[width=0.98\textwidth]{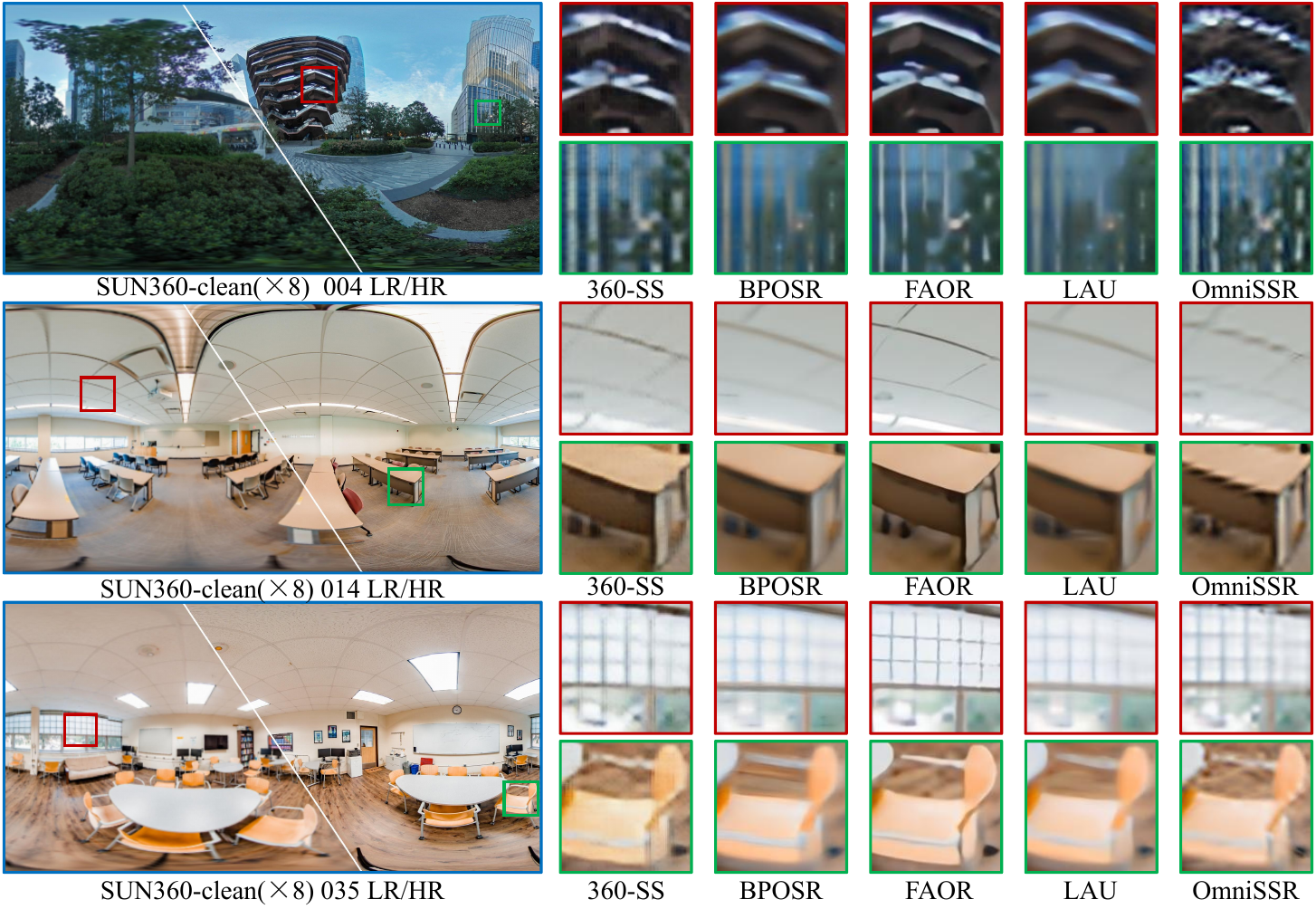}  
    \caption{Visual comparison of the super-resolution results of 360-SS~\cite{360SS}, BPOSR~\cite{BPOSR}, FAOR~\cite{FAOR}, LAU~\cite{lau}, and OmniSSR~\cite{omnissr} on the SUN360-clean~\cite{OSRT} dataset at $\times$8.}  
    \label{fig:x8sun}  
\end{figure*}

\begin{figure*}[!h]  
    \centering  
    \includegraphics[width=0.98\textwidth]{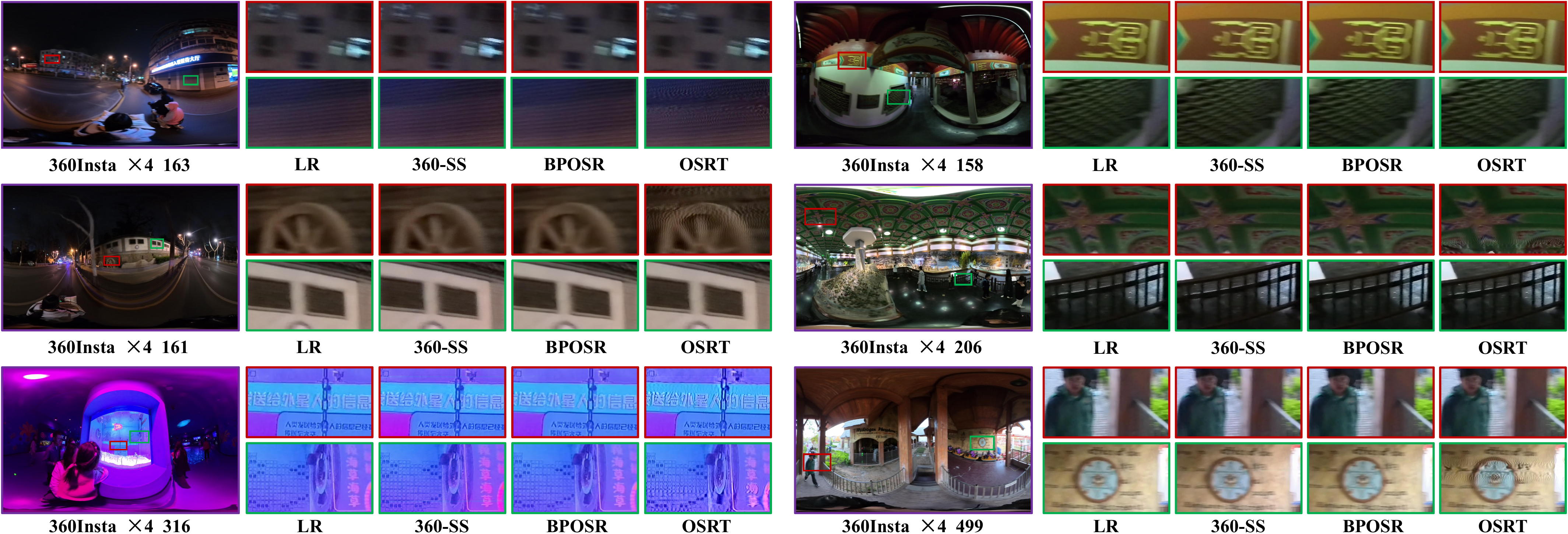}  
    \caption{Visual comparison of the super-resolution results of 360-SS~\cite{360SS}, BPOSR~\cite{BPOSR}, and OSRT~\cite{OSRT} on the 360Insta dataset at $\times$4.}  
    \label{fig:x4insta}  
\end{figure*}  



\section{Benchmarking and Empirical Analysis}
\label{sec:evaluation}
\subsection{Evaluation Metrics}  
To better assess the results of the ODISR and ODVSR tasks, image quality assessment (IQA) metrics are significant. Current evaluations are divided into objective and subjective assessments. 
Besides subjective assessments, objective assessments, including both full-reference and non-reference IQA, can objectively evaluate image quality. Additionally, in evaluating and optimizing the performance and efficiency of neural networks, it is also crucial to assess the model parameters (Params) and the number of floating-point operations (FLOPs) in deep learning models. We introduce these evaluation metrics below. 

\subsubsection{Full-Reference Metrics}  
\textbf{PSNR and WS-PSNR.} PSNR~\cite{assess} and WS-PSNR~\cite{WSPSNR} are currently the most widely used IQA metrics in the field of ODISR and ODVSR. The values closer to infinity are better. PSNR is widely used due to its simplicity, while WS-PSNR, by incorporating geometric distortion weights, aligns better with human visual perception and is especially suitable for quality assessment of wide-angle and spherical content.
\textbf{SSIM and WS-SSIM.} SSIM~\cite{assess} and WS-SSIM~\cite{WSSSIM} are used to measure the similarity between two images. The SSIM evaluation considers the image structure information based on perception. Compared to traditional SSIM, WS-SSIM incorporates a weighted mechanism for geometric distortions, making it more suitable for the structural characteristics of wide-angle and spherical images.

\subsubsection{Non-Reference Metrics}  
Non-Reference Image Quality Assessment (NR-IQA) is an advanced image quality evaluation method that does not require any reference images at all, but estimates image quality based on the characteristics of the distorted image itself. In recent years, a variety of NR-IQA methods have emerged, for example, \textbf{NIQE}~\cite{niqe} is suitable for image quality assessment based on natural scene statistics; \textbf{MUSIQ}~\cite{musiq} provides accurate evaluation results through multi-scale and self-supervised learning; \textbf{MANIQA}~\cite{maniqa} performs well in evaluating the quality of GAN generated images; \textbf{CLIP-IQA}~\cite{clipiqa} uses semantic understanding capabilities to provide high-quality evaluation.

There are some identified issues with relying solely on WS-PSNR/WS-SSIM and PSNR/SSIM as evaluation metrics. Although these metrics have reached a level of effectiveness, images reconstructed using them tend to become overly smooth and blurry, which does not align with human visual perception.  
In response, researchers have started to emphasize the importance of perceptual quality and have proposed subjective image quality evaluation metrics. However, there is still no universally accepted metric in the field. This highlights a critical area where advances are needed in the current landscape of research.

\subsection{Experimental Results and Analysis}  
This section provides empirical analysis and highlights some key challenges in deep learning-based ODISR and ODVSR. We conduct extensive evaluations on several benchmarks (e.g. ODISR~\cite{lau}, ODISR-clean~\cite{OSRT}, SUN360~\cite{lau}, and SUN360-clean~\cite{OSRT}) and our proposed 360Insta dataset. In the experiments, we compare the deep learning-based methods, including 360-SS~\cite{360SS}, OSRT~\cite{OSRT}, OSRT-Light~\cite{OSRT}, BPOSR~\cite{BPOSR}, LAU~\cite{lau}, and OmniSSR~\cite{omnissr} methods. For all compared methods, we use the publicly available code to produce their results for fair comparisons, aiming to facilitate future researchers in comparison and evaluation.


\subsubsection{ODISR Performance Comparison under ERP Downsampling Setting}  
Following the testing protocol established by LAU~\cite{lau}, we conduct SR tests for LR images using deep learning-based ODISR methods, specifically for scaling factors of 2, 4, 8, and 16, on the ODISR~\cite{lau} and SUN360~\cite{lau} datasets. LR images $LR_{odi}$ are generated through bicubic downsampling of HR images $HR_{odi}$, with the test results presented in Table~\ref{tab:ori_odisr_results}.  

As shown in Table~\ref{tab:ori_odisr_results}, the performance of each method at different scales varies significantly.
At a scaling factor of 2, the 360-SS\cite{360SS} method achieved the best performance of PSNR and SSIM, while WS-PSNR and WS-SSIM do not perform well on the ODISR dataset~\cite{lau}, which demonstrates that 360-SS\cite{360SS} performs poorly in applications that emphasize details.
At a scaling factor of 4 and 8, the BPOSR~\cite{BPOSR} method achieves the best performance on the ODISR dataset~\cite{lau}. Its superior performance can be attributed to the innovative use of ERP and CMP projections, which effectively enhance the model's ability to learn the geometric structure of ODIs. OmniSSR~\cite{omnissr} demonstrates excellent results, benefiting from its sectional projection design that effectively reduces the gap between ISR and ODISR. This approach helps to better mitigate distortions and artifacts specific to panoramic imagery, leading to robust testing performance.
At a scaling factor of 16, the BPOSR~\cite{BPOSR} method performs well in PSNR and SSIM, reaching 24.01 and 0.6730, respectively, but is slightly inferior in WS-PSNR and WS-SSIM, indicating that it still needs to be improved in subjective perception.

\subsubsection{ODISR Performance Comparison under Fisheye Downsampling Setting}  
Based on the testing protocol established by OSRT~\cite{OSRT}, we conduct SR tests for LR images using deep learning-based ODISR methods on the ODISR-clean~\cite{OSRT} and SUN360-clean~\cite{OSRT} datasets, specifically for scaling factors of 2, 4, 8, and 16. Data pairs are generated through fisheye downsampling, followed by data adjustment through cleaning. The test results are presented in Table~\ref{tab:fish_odisr_results}.  

As shown in Table~\ref{tab:fish_odisr_results}, OSRT~\cite{OSRT} outperforms other methods at both $\times$2 and $\times$4 scaling factors for the ODISR task. This superiority primarily stems from its incorporation of a fisheye-aware downsampling mechanism and extensive training tailored to this specific distortion type. Although BPOSR~\cite{BPOSR} demonstrates robust performance across various scenarios, it still falls short of OSRT~\cite{OSRT} in this particular context. Conversely, OmniSSR~\cite{omnissr} exhibits inferior performance on this distorted fisheye dataset, largely due to its lack of dedicated mechanisms for fisheye distortion adaptation and insufficient training on such aberrations. 
These findings highlight the importance of designing specialized models that explicitly address unique distortion characteristics, especially for distortion scenarios in ODISR tasks.  

The above experimental comparisons indicate that as the scaling factor increases, the degree of performance improvement of the methods tends to diminish, highlighting the challenges still faced in SR tasks for large scaling factors. Furthermore, while the OSRT~\cite{OSRT} method shows considerable advantages under its data adjustment testing protocol, its performance declines sharply under the original ERP downsampling, suggesting that both data cleaning and downsampling possess effectiveness.

\subsubsection{Model Complexity Comparison} 
 As shown in Table~\ref{tab:param}, at a scaling factor of 4, we compare the computational complexity of ODISR methods, including FLOPs and Params averaged over 100 ODIs of size 512$\times$256 using an NVIDIA 3090Ti GPU. 
 As can be seen from Table~\ref{tab:param}, the number of parameters of OSRT~\cite{OSRT} is 14.2M, indicating that the model has high complexity in feature extraction and may provide better results for ODISR. This is because OSRT~\cite{OSRT} designs the complex attention mechanism based on Swin-Transformer~\cite{swinsr}. In contrast, 360-SS~\cite{360SS} has the lowest number of parameters, only 1.6M, suggesting that it uses a lighter structure and is suitable for resource-limited application scenarios, but its performance is lower. This is because the architecture used is relatively simple, comprising fewer layers and parameters, with lightweight residual blocks and a PatchGAN discriminator.

\subsubsection{ODVSR Performance Comparison}
Since the code of current ODVSR methods is not fully accessible, we use deep learning-based VSR methods to perform evaluations on the MiG Panorama~\cite{SMFN} and 360VDS~\cite{S3PO} datasets. The test results are shown in Table~\ref{tab:fish_odvsr_results}.

As shown in Table~\ref{tab:fish_odvsr_results}, at the scaling factor of 4, in the MiG Panorama dataset~\cite{SMFN}, SAVSR shows good performance with WS-PSNR and WS-SSIM of 29.12dB and 0.8227, respectively. On the 360VDS dataset, the MIAVSR method shows a good performance of 24.47dB and 0.7669, respectively.
These results provide a performance reference for traditional VSR algorithms in the field of ODVSR.

\subsubsection{ODISR Performance Comparison on 360Insta Dataset}
While existing models have demonstrated remarkable improvements in handling bicubic degradation data, the types of image degradation encountered in real-world applications are complex and diverse. Phenomena such as motion blur, low illumination, and other degradation issues are prevalent. These intricate degradation factors are highly likely to undermine the effectiveness of SR tasks. Consequently, relying solely on the performance of synthetic degradation data is insufficient to comprehensively evaluate the robustness and practicality of a model. To address this, this section employs our newly proposed 360Insta dataset and utilizes non-reference metrics to evaluate the quality of images restored by different methods. This approach provides a multi-faceted reflection of the model's performance in real-world degradation scenarios, offering a more representative basis for the practical promotion of SR models.

Specifically, due to computational resource limitations, we divide 360Insta images into 16 blocks to evaluate the performance of $\times$2 and $\times$4 SR models. As shown in Table~\ref{tab:360insta}, we employed four non-reference image quality assessment metrics, including NIQE, MUSIQ, MANIQA, and CLIP-IQA, to objectively measure the restoration quality. 

\begin{table}[h] 
\centering 
\caption{Comparison of Deep Learning-based ODISR Methods on 360Insta Dataset. NIQE $\downarrow$: the lower, the better; MUSIQ/MANIQA/CLIPIQA $\uparrow$: the higher, the better. The best result is in {\color{red}{red}}, whereas the second and third best results are in {\color{blue}{blue}} and {\color{green}{green}} under each case, respectively.} 
\label{tab:360insta} 
\resizebox{0.5\textwidth}{!}{ 
\begin{tabular}{c|c|c|c|c|c}  
\hline  
Method & Scale & NIQE $\downarrow$ & MUSIQ $\uparrow$ & MANIQA $\uparrow$& CLIPIQA $\uparrow$\\
\hline  
360-SS~\cite{360SS} &$\times$2  & {\color{blue}6.7448} & {\color{red}32.109} & {\color{blue}0.2553} & {\color{blue}0.2704} \\
OSRT~\cite{OSRT} & $\times$2 & {\color{red}5.0306}  & {\color{blue}29.427}  & {\color{red}0.2878}  & {\color{red}0.4186} \\
\hline   
\hline   
360-SS~\cite{360SS} & $\times$4 & {\color{blue}8.5964} & {\color{blue}22.188} & {\color{green}0.2819}& {\color{green}0.3148}  \\
OSRT~\cite{OSRT} &$\times$4  &{\color{red}8.1042} & {\color{red}24.334} & {\color{blue}0.3019} &{\color{red}0.4763}  \\
BPOSR~\cite{BPOSR} &$\times$4  &{\color{green}10.6850} &{\color{green}22.048}  & {\color{red}0.3084} & {\color{blue}0.4505} \\
\hline  
\end{tabular}  
}  

\end{table}  

As shown in Table~\ref{tab:360insta}, in scaling factor of $\times$2, the NIQE~\cite{niqe} of the OSRT method is 5.0306, which performs the best, while the NIQE~\cite{niqe} of 360-SS~\cite{360SS} is 6.7448, which is relatively higher. This indicates that the OSRT~\cite{OSRT} method is more effective in maintaining image quality. As the scaling factor increases to $\times$4, the NIQE~\cite{niqe} for all methods generally rises, reflecting a trend of deterioration in image quality assessed through non-reference metrics at higher scales. Overall, the non-reference image quality assessment metrics are low, indicating that the current model does not perform well on the real degradation dataset.

\subsubsection{Visual Quality Comparison}  
In this section, we provide a visual comparison between ODISR performance on the ODISR-clean dataset~\cite{OSRT}, the SUN360-clean datasets~\cite{OSRT}, and our proposed 360Insta dataset.  
As illustrated in Figs.~\ref{fig:x2odisr} and~\ref{fig:x2sun}, we present visual results of the ODISR methods for 2$\times$ on the ODISR~\cite{OSRT} and SUN360-clean~\cite{OSRT} datasets. Notably, 360-SS~\cite{360SS} exhibits significant noise artifacts, while BPOSR~\cite{BPOSR} produces overly smoothed results. Omnissr~\cite{omnissr} shows good restoration results in areas with fewer textures, because in these areas, the details are simpler and change slowly, and the model can effectively use contextual information for smoothing, thus generating clear image results. However, in regions with more complex textures, the Omnissr~\cite{omnissr} method exhibits relatively inferior performance. This is mainly due to the presence of complex textures, which leads to excessive smoothing in the image. In areas with high-frequency details, the model may not be able to accurately capture the richness of the texture and the changes in details, and thus cannot effectively reconstruct the features of the original image. In contrast, OSRT~\cite{OSRT} achieves superior restoration in terms of human visual perception.
The challenges are compounded when increasing the upscaling factor to 4$\times$ and 8$\times$, as shown in Figs.~\ref{fig:x4odisr}-\ref{fig:x4sun} and Figs.~\ref{fig:x8odisr}-\ref{fig:x8sun}. The resulting image quality is correspondingly reduced. Furthermore, LAU~\cite{lau} introduces noticeable discontinuities between latitudinal bands, negatively affecting the overall perceptual quality.

The above experimental comparisons indicate that OSRT~\cite{OSRT} reveals pronounced artifacts that compromise visual coherence while it still performs well. Additionally, LAU~\cite{lau} exacerbates the problem by producing stark discontinuities along latitudinal bands, further diminishing the overall perceptual quality. These findings underscore the complexities associated with ODISR and ODVSR, particularly at larger scaling factors. Therefore, current research on ODIs and ODVs still has considerable room for improvement, particularly in the design of datasets and methodologies.

As illustrated in Fig.~\ref{fig:x4insta}, the image restoration effect of the current methods on our 360Insta dataset is not ideal, especially in dim environments and motion blur conditions, where the restoration performance is significantly limited. Therefore, in order to comprehensively evaluate and further improve the robustness of the model, it is urgent to introduce more representative and diverse data for systematic testing. Our 360Insta dataset can play an important role in assessing and improving model performance, helping the current model achieve more accurate and robust restoration effects. In addition, we also provide a 360Insta video dataset for testing the ODVSR task~\url{https://drive.google.com/drive/my-drive}. Due to space constraints, the 360Insta video dataset is not included in this paper.
\section{Conclusion and Future Work}  
With the advancement of deep learning, ODISR models have demonstrated excellent performance on synthetic bicubic degradation data. In terms of visual quality, quantitative and qualitative analyses reveal that OSRT~\cite{OSRT} exhibits strong detail recovery capabilities, while BPOSR~\cite{BPOSR} demonstrates robust performance but tends to produce over-smoothing artifacts. 360-SS~\cite{360SS} offers fast inference but is prone to introducing noise artifacts, which affect visual experience. As the SR magnification increases, model performance gradually declines, and the recovery of fine details becomes increasingly challenging, exposing limitations in SR capacity. In terms of model complexity, for instance, SwinIR~\cite{swinsr} and OSRT~\cite{OSRT} possess large parameter counts and strong expressive power but demand substantial hardware resources for training and inference. Conversely, lightweight models such as 360-SS~\cite{360SS} have fewer parameters and lower computational costs, making them suitable for resource-constrained scenarios; however, their performance is relatively inferior, struggling to meet HR restoration requirements. Consequently, high-precision models come with significant computational costs, highlighting the trade-off between performance and complexity. 
The current ODISR method has good recovery effects in the synthesis of bicubic degradation. However, their robustness is limited when confronted with real-world degradations, such as motion blur, low-light conditions, and underexposure, across diverse scenes and different scales. Consequently, there is an urgent need to introduce more real-world evaluation datasets that better reflect practical degradations. 
Overall, achieving more robust and efficient ODISR remains a challenge that calls for further exploration in model design, data diversity, and the balance between accuracy and efficiency to better address the demands of real-world scenarios.

Deep learning-based SR methods still face several challenges, which represent future research trends.  

\noindent
\textbf{Network Architectures}: The design of network architecture plays a crucial role in determining the enhancement performance. As previously noted, most deep models for ODISR leverage CNNs, Transformers, or hybrid architectures. Although these approaches have achieved promising results in ODISR tasks, challenges remain, particularly regarding computational complexity and geometric distortions in models.
Some architectures, due to their large parameter counts, can effectively recover fine textures; however, this often comes at the cost of increased memory consumption and longer inference times. Others incorporate projection strategies aimed at mitigating ERP distortion but tend to introduce additional computational overhead.
Therefore, developing more efficient ODISR architectures is imperative, especially considering the unique characteristics of panoramic imagery, such as a wide field of view and geometric distortions. Given the complex spatial relationships among pixels in ODIs, the multimodal properties of models like Sora~\cite{Sora} can enhance understanding of spatial occupancy, thereby better addressing geometric distortions and data sparsity issues. By integrating these advanced techniques, researchers can further explore how to incorporate prior knowledge of image properties into network design, ultimately improving SR performance.

\noindent
\textbf{ODISR for Arbitrary Scaling Factors}: 
Current ODISR methods are typically trained for fixed scaling factors, and as the scaling factor increases, the performance of the networks often declines significantly, resulting in a lack of effective adaptability for other scales. Considering the unique challenges posed by ODIs in ERP projection, such as severe deformation in high-latitude regions, rich textures, and complex structures, fixed-scale ODISR methods struggle to restore details and compensate for shape distortions across all regions. Thus, developing ODISR models capable of handling arbitrary scaling factors through multi-scale feature extraction, adaptive interpolation strategies, generative adversarial networks, and diverse dataset training is an essential research direction. This approach is critical to meet the practical requirements for high-quality ODISR reconstruction of ODIs and ODVs, especially in dealing with the complex characteristics of ODIs, and will significantly improve the adaptability and effectiveness of ODISR techniques in real-world applications.

\noindent
\textbf{360° Data}: The current landscape of ODIs still faces significant challenges due to the scarcity of diverse, real-world paired datasets. Most existing datasets rely on simplified synthetic degradations, such as bicubic downsampling, which fail to accurately model the complex degradations encountered during actual display processes. To advance the field, it is crucial to formulate more realistic degradation models and develop large-scale, diverse datasets capturing ODIs under various display conditions. Future efforts should focus on multi-source and multi-modal data collection, such as leveraging mobile devices to capture real-world scenes under different display scenarios or employing sophisticated synthesis techniques to generate more photorealistic degraded samples. These initiatives are essential to drive the next wave of progress in panoramic super-resolution and expand the applicability of this technology in real-world applications.

\textbf{Extension to ODVSR}: Despite the rapid progress in traditional ISR~\cite{seesr} and VSR~\cite{upscale} methods, the research on ODVSR remains relatively underexplored and develops at a slow pace. Major challenges include inter-frame misalignment and artifact generation, particularly when existing video SR techniques are directly applied to high-dimensional panoramic videos, often resulting in unacceptable artifacts that severely degrade visual quality. Addressing these issues requires the development of specialized SR algorithms tailored to the unique characteristics of ODVs, with a focus on mitigating artifacts and improving temporal consistency, thereby accelerating the advancement and deployment of ODVSR.

The SR reconstruction of ODIs and ODVs is a research field filled with challenges and opportunities. Although current technologies have made significant progress, there is still a need to continuously explore new methods and techniques to address increasingly complex visual tasks. Through in-depth research on network architectures, loss functions, datasets, and evaluation metrics, future studies will provide more effective solutions for the SR reconstruction of ODIs and ODVs, driving further advancement in this field.

	{
		\bibliographystyle{IEEEtran}
		\bibliography{360_bib}
	}

\end{document}